\documentclass[11pt]{article}
\usepackage[T1]{fontenc}
\usepackage[utf8]{inputenc}
\usepackage{lmodern}
\usepackage[a4paper,margin=2.5cm]{geometry}
\usepackage{microtype}
\usepackage{amsmath,amssymb}
\usepackage{booktabs}
\usepackage{longtable}
\usepackage{graphicx}
\usepackage{xcolor}
\usepackage{adjustbox}
\usepackage{float}
\usepackage{fvextra}
\usepackage{enumitem}
\usepackage{caption}
\usepackage[numbers,sort&compress]{natbib}
\usepackage[colorlinks=true,linkcolor=blue!50!black,citecolor=blue!50!black,urlcolor=blue!50!black]{hyperref}
\usepackage{xurl}

\newcommand{\zenodoDOI}{10.5281/zenodo.24375031}
\newcommand{\repoURL}{https://github.com/agumus/EvidenceAgentBench}
\newcommand{\passk}{pass$^3$}
\newcommand{\versionDate}{27 November 2026}  %
\newcommand{\sampleManifestHash}{98a84902\allowbreak a731db7b\allowbreak 88350fa2\allowbreak fac9af57\allowbreak 512aad88\allowbreak fcd10439\allowbreak 1c7be8e8\allowbreak a7d78ab6}  %
\newcommand{\nSampleItems}{13}  %
\newcommand{\nCandidates}{85}  %
\newcommand{\candidateQueryEnd}{2026/10/21}
\newcommand{\nExcluded}{35}

\newcommand{\nReviews}{50}
\newcommand{\nItems}{65}
\newcommand{\nSecondQ}{15}

\newcommand{\nDirNote}{15}
\newcommand{\nCleanItems}{40}
\newcommand{\nControls}{10}
\newcommand{\nUpdates}{29}
\newcommand{\firstPublishedMin}{2 July 2026}
\newcommand{\firstPublishedMax}{13 October 2026}

\newcommand{\nPrevSame}{18}
\newcommand{\nPrevChanged}{11}
\newcommand{\nPrevNotComparable}{5}
\newcommand{\nPrevNone}{31}
\newcommand{\exclusionList}{protocol 13; surrogate 10; ambiguous 2; overview 2; reissue 2; no estimate 1; scoping 1; qualitative 1; methodology 1; non patient outcome 1; diagnostic 1}

\newcommand{\nOpenRollouts}{780}
\newcommand{\nOpenLimitHits}{11}

\newcommand{\openMeanMin}{0.839}
\newcommand{\openMeanMax}{0.895}
\newcommand{\passThreeMin}{0.71}
\newcommand{\passThreeMax}{0.89}
\newcommand{\worstGapMin}{0.09}
\newcommand{\worstGapMax}{0.15}
\newcommand{\topThreeMeanSpread}{0.02}  %
\newcommand{\topThreePassSpread}{0.09}  %
\newcommand{\pOneOpen}{8}  %
\newcommand{\mdd}{0.07}
\newcommand{\nBoot}{1{,}000}
\newcommand{\largestPairA}{muse-glimmer}
\newcommand{\largestPairB}{ornith-1.5}
\newcommand{\largestPairDiff}{+0.056}
\newcommand{\largestPairCI}{[$-$0.000, 0.119]}

\newcommand{\cFourLostTotal}{213}
\newcommand{\cFourWithCThree}{92\%}
\newcommand{\cFourLostNever}{11}  %
\newcommand{\cFourLossHigh}{10\%}
\newcommand{\cFourLossModerate}{29\%}
\newcommand{\cFourLossLow}{48\%}
\newcommand{\cFourLossVeryLow}{5\%}
\newcommand{\lossCOneMin}{3\%}
\newcommand{\lossCOneMax}{7\%}
\newcommand{\lossCTwoMin}{13\%}
\newcommand{\lossCTwoMax}{17\%}
\newcommand{\lossCThreeMin}{2\%}
\newcommand{\lossCThreeMax}{4\%}
\newcommand{\lossCFourMin}{19\%}
\newcommand{\lossCFourMax}{33\%}
\newcommand{\lossCSevenMin}{11\%}
\newcommand{\lossCSevenMax}{21\%}
\newcommand{\reviewRetrievedMin}{83\%}
\newcommand{\reviewRetrievedMax}{87\%}
\newcommand{\citedPmidShareMin}{3.9\%}
\newcommand{\citedPmidShareMax}{5.8\%}
\newcommand{\cleanSubmittedMin}{0.56}
\newcommand{\cleanSubmittedMax}{0.64}
\newcommand{\CleanNemotron}{0.570}

\newcommand{\CleanSubNemotron}{39}

\newcommand{\CleanOrnith}{0.320}

\newcommand{\CleanSubOrnith}{20}
\newcommand{\nCleanRollouts}{160}
\newcommand{\nCleanCapHits}{43}
\newcommand{\cleanReviewRetrieved}{0}

\newcommand{\BudgetMuseNew}{0.569}
\newcommand{\BudgetMuseDiff}{+0.295}
\newcommand{\BudgetMuseCI}{[+0.163, +0.410]}

\newcommand{\BudgetQwenDiff}{+0.072}
\newcommand{\BudgetQwenCI}{[$-$0.094, +0.238]}

\newcommand{\BudgetOrnithDiff}{+0.286}
\newcommand{\BudgetOrnithCI}{[+0.167, +0.416]}
\newcommand{\BudgetOrnithCapNew}{7}
\newcommand{\budgetItems}{24}
\newcommand{\budgetUsedMore}{39}
\newcommand{\budgetGainAbout}{0.29}  %

\newcommand{\budgetRollouts}{72}
\newcommand{\budgetCut}{42}
\newcommand{\budgetCutFitBase}{15}
\newcommand{\budgetCutUsedMore}{20}
\newcommand{\budgetCutAgain}{7}
\newcommand{\budgetLongestK}{69}
\newcommand{\NemotronCleanAffected}{0.562}
\newcommand{\openHoursTotal}{26}
\newcommand{\cleanHoursTotal}{9}
\newcommand{\nWithCFive}{34}
\newcommand{\nWithCSeven}{61}
\newcommand{\posPointsMin}{10}
\newcommand{\posPointsMax}{12}
\newcommand{\prevGapMin}{0.03}
\newcommand{\prevGapMax}{0.15}
\newcommand{\prevGapOrnith}{0.146}
\newcommand{\ctrlOpenRollouts}{120}
\newcommand{\ctrlOpenRefusals}{0}
\newcommand{\pFiveOtherOpen}{1}
\newcommand{\ctrlCleanRollouts}{12}
\newcommand{\ctrlCleanRefusals}{0}
\newcommand{\pFiveOtherClean}{8}
\newcommand{\ftSample}{60}
\newcommand{\ftPerAgent}{15}

\newcommand{\ftFailedOpen}{20}
\newcommand{\ftFailedClean}{17}
\newcommand{\ftOutdatedOpen}{17}
\newcommand{\ftHalluOpen}{3}
\newcommand{\ftUnderClean}{10}
\newcommand{\ftOverClean}{2}
\newcommand{\ftUnderCleanAll}{11}

\newcommand{\ftHarmsAddressed}{60}
\newcommand{\ftSevenLost}{38}
\newcommand{\ftAnsweredOpen}{769}

\newcommand{\ftFailOpenAll}{51}

\newcommand{\ftNeverOpen}{11}
\newcommand{\ftNeverClean}{43}
\newcommand{\ftNeverSearchClean}{22}
\newcommand{\ftNeverFetchClean}{3}
\newcommand{\ftNeverDistinct}{99.5\%}
\newcommand{\accReadN}{639}
\newcommand{\accReadFailed}{15}
\newcommand{\accReadPct}{2.3\%}
\newcommand{\accReadMean}{0.930}

\newcommand{\accNeverN}{114}

\newcommand{\accNeverPct}{29.8\%}
\newcommand{\accNeverMean}{0.663}
\newcommand{\accNotReadFailed}{36}
\newcommand{\accNotReadN}{130}
\newcommand{\accPenaltyFailed}{15}
\newcommand{\accPassNoPenalty}{15}
\newcommand{\accWithinN}{42}
\newcommand{\accWithinDiff}{0.317}
\newcommand{\accWithinLo}{0.253}
\newcommand{\accWithinHi}{0.377}
\newcommand{\cfN}{284}
\newcommand{\cfS}{202}
\newcommand{\cfSPct}{71\%}
\newcommand{\cfW}{55}
\newcommand{\cfWPct}{19\%}
\newcommand{\cfHM}{26}
\newcommand{\cfHMPct}{9\%}

\newcommand{\cfOpenSPct}{78\%}
\newcommand{\cfOpenW}{30}

\newcommand{\cfCleanSPct}{55\%}

\newcommand{\cfReadS}{114}
\newcommand{\cfReadSPct}{86\%}
\newcommand{\cfReadSOfAll}{18\%}
\newcommand{\cfNotReadW}{21}
\newcommand{\cfSNoDiff}{92}
\newcommand{\cfSOther}{110}
\newcommand{\cfOverlapAgree}{35}
\newcommand{\cfOverlapBoth}{37}
\newcommand{\ptFired}{29}

\newcommand{\ptRescued}{12}
\newcommand{\ptRescuedOrnith}{7}
\newcommand{\ptDisclosedJudge}{18}
\newcommand{\ptOrnithWith}{0.839}
\newcommand{\ptOrnithWithout}{0.866}
\newcommand{\ptSampleDisc}{5}
\newcommand{\ptSampleDiscAll}{28}
\newcommand{\ptSampleNone}{31}
\newcommand{\poN}{10}
\newcommand{\poSlips}{8}
\newcommand{\poSlipExist}{2}
\newcommand{\poFailed}{10}
\newcommand{\poRescued}{9}

\newcommand{\jcN}{60}
\newcommand{\jcOneAgree}{53}
\newcommand{\jcOneLenient}{6}
\newcommand{\jcFourAgree}{53}
\newcommand{\ftSpotN}{10}
\newcommand{\ftSpotAgreed}{10}
\newcommand{\relN}{20}
\newcommand{\relPrimaryAgree}{12}
\newcommand{\relPrimary}{0.52}
\newcommand{\relMergedAgree}{17}
\newcommand{\relMerged}{0.81}
\newcommand{\relReversed}{5}
\newcommand{\relDirectionAgree}{20}
\newcommand{\relWording}{0.40}
\newcommand{\relIndirect}{0.39}
\newcommand{\exScoreOutdated}{0.273}
\newcommand{\exScoreOverclaim}{0.750}
\newcommand{\jtwoTag}{\texttt{laguna-s-2.1:dflash-nvfp4}}
\newcommand{\jtwoThreeTag}{\texttt{ripple-preview:65b}}

\newcommand{\jtwoSampleAnswers}{52}
\newcommand{\jtwoCleanAnswered}{117}

\newcommand{\jtwoDecisions}{1{,}772}
\newcommand{\jtwoReviews}{38}
\newcommand{\jtwoAgree}{0.898}
\newcommand{\jtwoAgreeCI}{[0.882, 0.915]}
\newcommand{\jtwoKappa}{0.793}
\newcommand{\jtwoKappaCI}{[0.758, 0.827]}
\newcommand{\jtwoACone}{0.801}

\newcommand{\jtwoDirSecond}{131}  %
\newcommand{\jtwoDirRecord}{49}  %
\newcommand{\jtwoPtwoRecord}{1\%}
\newcommand{\jtwoPtwoSecond}{43\%}

\newcommand{\jtwoHarmsRecord}{43\%}
\newcommand{\jtwoHarmsSecond}{31\%}

\newcommand{\jtwoNoPtwoAgree}{0.932}
\newcommand{\jtwoNoPtwoAgreeCI}{[0.913, 0.949]}
\newcommand{\jtwoNoPtwoKappa}{0.863}

\newcommand{\jtwoOpenAnswers}{769}

\newcommand{\jtwoOpenDecisions}{8{,}047}
\newcommand{\jtwoOpenAgree}{0.941}
\newcommand{\jtwoOpenAgreePct}{94\%}
\newcommand{\jtwoOpenAgreeCI}{[0.933, 0.949]}
\newcommand{\jtwoOpenKappa}{0.879}

\newcommand{\jtwoOpenNoPtwoAgree}{0.962}
\newcommand{\jtwoOpenCFourLossRecord}{26.3\%}
\newcommand{\jtwoOpenCFourLossSecond}{27.7\%}

\newcommand{\jtwoCleanRankedRecord}{4}
\newcommand{\jtwoCleanRankedSecond}{2}

\newcommand{\jtwoCleanDropMin}{0.05}  %
\newcommand{\jtwoCleanDropMax}{0.16}

\newcommand{\jtwoOpenRankedSecond}{3}

\newcommand{\jtwoOpenDropMin}{0.04}  %
\newcommand{\jtwoOpenDropMax}{0.10}
\newcommand{\jtwoThreeAgree}{0.945}
\newcommand{\jtwoThreeKappa}{0.884}

\newcommand{\jtwoThreePtwo}{5\%}  %
\newcommand{\jtwoFleiss}{0.828}
\newcommand{\jtwoFleissCI}{[0.800, 0.855]}
\newcommand{\jtwoUnanimous}{87\%}
\newcommand{\jtwoHarmsUnanimous}{78\%}  %

\newcommand{\nEmptyCheck}{745}

\title{EvidenceAgentBench: rubric-scored evaluation of LLM agents on clinical evidence synthesis%
\thanks{Version 1, \versionDate{}. All results use author-drafted (draft-0) rubrics and local LLM judges. The physician review of the rubrics is under way and the blinded physician grading is planned, under a protocol fixed in this version; no physician decision enters this version. Both will be reported in a revised version.}}
\author{Abdurrahman Gumus\\
MIRA Lab, Department of Computer Engineering\\
Isparta University of Applied Sciences, Isparta, T\"urkiye\\
\texttt{abdurrahmangumus@isparta.edu.tr}}
\date{\versionDate{}}

\begin{document}
\maketitle

\begin{abstract}
Clinicians increasingly ask language models what the evidence says about a treatment, and agents that search the literature are now practical. Whether their answers match the best available synthesis, and state its certainty correctly, is rarely measured. EvidenceAgentBench is a benchmark of \nItems{} clinical PICO questions from \nReviews{} Cochrane systematic reviews published between July and October 2026, newer than the published training cut-offs of major commercial models. An agent searches PubMed and Europe PMC and submits a structured answer, which an LLM judge grades criterion by criterion against an item-specific rubric: direction, estimate, GRADE certainty and its wording, harms, and penalties such as unretrieved citations and overclaiming. All inference and grading run locally. For four open-weight agents (\nOpenRollouts{} Open and \nCleanRollouts{} Clean-room rollouts), mean Open scores, where the review can be found, were \openMeanMin{} to \openMeanMax{}; no pair of agents could be ranked, and certainty-matched wording was the most frequent loss (\lossCFourMin{} to \lossCFourMax{} of rollouts), worded too strongly in \cfOpenSPct{} of those answers. Answers that read the review failed in \accReadPct{} of cases, against \accNeverPct{} of those that never retrieved it. With the review hidden and the tools stopped at its search date, scores fell to \CleanOrnith{} to \CleanNemotron{}, and a run with twice the message budget showed that the Clean-room ranking depends on the budget. A second judge from another model family agreed with the first on \jtwoOpenAgreePct{} of Open criterion decisions and left the wording result unchanged; one penalty, for citing an irrelevant source, carried most of the disagreement and moved agent rankings. The rubrics are author-drafted: their physician review is under way and blinded physician grading of 52 answers frozen in advance is planned, under a pre-specified protocol; both will be reported in a revised version. Code, items, rubrics and most transcripts are openly released.
\end{abstract}

\section{Introduction}

Clinicians increasingly ask language models what the evidence says about a treatment, and tool-using agents can now run their own literature searches before they answer. Such an answer is useful only if it reproduces what a careful synthesis of the evidence would conclude: the direction of the effect, its size with its uncertainty, how certain the evidence is, what limits that certainty, and what is known about harms. It is harmful if it cites sources that do not exist, presents low-certainty evidence as settled, or turns a population-level summary into advice for an individual patient. Systematic reviews and the GRADE approach~\citep{guyatt2008grade,higgins2024handbook} exist to make exactly these judgements explicit, which makes them a natural reference for evaluating evidence-synthesis agents.

Existing benchmarks cover parts of this task. Rubric-graded benchmarks such as HealthBench~\citep{arora2025healthbench} grade conversational replies against physician-written criteria; agentic benchmarks such as MedAgentBench~\citep{jiang2025medagentbench} evaluate actions in an electronic health record; and evidence-synthesis benchmarks either give the model the source studies~\citep{polzak2025medevidence,ha2026medmeta}, use reviews published before current training cut-offs~\citep{jung2026sciconharness}, or evaluate the stages of producing a review rather than answering a clinical question~\citep{xie2026metasyn,zabaleta2026scilitbench}. A systematic review of LLM evaluation in health care found that only 1.2\% of studies assessed calibration or uncertainty~\citep{bedi2025jama}, and where LLM judges have been compared with clinicians, their agreement was close to the agreement between clinicians themselves~\citep{arora2025healthbench,bedi2025medhelm}, which makes validation on the task at hand necessary.

We ask three questions: how well locally run agents reproduce a recent review's conclusion, certainty and harms when they can find it; how they do when they cannot; and why they fail.

EvidenceAgentBench combines these elements in one small, transparent benchmark. The agent is not given the studies: it answers a clinical PICO question by searching PubMed and Europe PMC and reading the abstracts itself, and its answer is graded against the conclusion, the certainty and the harms of a recent Cochrane review. Its contributions are:
\begin{itemize}[leftmargin=*,itemsep=2pt]
\item \textbf{Items with recent reference conclusions.} \nItems{} PICO questions from \nReviews{} Cochrane reviews first published between \firstPublishedMin{} and \firstPublishedMax{}, each with structured facts (direction, estimate, GRADE certainty, downgrade reasons, harms) and labels for the review's earlier version, for questions whose estimate and wording pull apart, and for over-refusal controls (Section~\ref{sec:construction}).
\item \textbf{GRADE-aware rubrics with penalties.} A rubric template with eight positive criteria and five penalties, instantiated per item and checked so that no criterion can be met by an empty answer (Section~\ref{sec:rubrics}). The rubrics in this version are the author's drafts; their physician review is under way, and no physician decision enters this version. The fabricated-citation penalty is computed from the tool trace, not judged.
\item \textbf{An agentic harness with a tool-trace audit} built on Inspect~\citep{inspect2024}, in two conditions: Open, and Clean-room with the tools stopped at the review's search date and the review hidden (Section~\ref{sec:harness}).
\item \textbf{A first evaluation} of four locally run open-weight agents, with review-cluster bootstrap intervals, \passk{} reliability, a retrieval audit, a pre-specified budget sensitivity analysis and a failure analysis of \ftSample{} coded transcripts (Section~\ref{sec:results}).
\item \textbf{Agreement between judges.} The same answers re-judged by a second local judge from another model family and, in part, by a third, with the effect of the judge on agent rankings (Section~\ref{sec:results}).
\item \textbf{A pre-specified physician-validation protocol.} Fifty-two answers were frozen with cryptographic hashes before any physician input; physicians will grade them blind to model and judge, and the analysis, including the rule that selects the primary judge, is fixed in this version (Section~\ref{sec:validation}). The results will be reported in a revised version.
\end{itemize}
The benchmark is local-first by design: every agent, the judge and all re-scoring run on one laptop, at no API cost, so that an evaluation of this kind can run where patient data are kept, for example inside a hospital. This study itself uses no patient data (Section~\ref{sec:localfirst}).

\section{Related work}
\label{sec:related}

\paragraph{Rubric-graded health benchmarks.} HealthBench~\citep{arora2025healthbench} grades replies to 5,000 health conversations against 48,562 physician-written criteria with positive and negative points, scored by a GPT-4.1 grader whose macro-F1 against physicians (0.709) was close to physician--physician agreement. We adopt its score formula, its negative points and its macro-F1 for judge validation. MedHELM~\citep{bedi2025medhelm} evaluates 121 clinician-validated tasks, with an LLM jury agreeing with clinicians at an intraclass correlation of 0.47 against 0.43 between clinicians. LiveMedBench~\citep{yan2026livemedbench} harvests new cases weekly to separate them in time from training data and grades them with automatically generated rubrics. Three industry benchmarks, reported on the company's web pages and combined in the micro1 Medical Index~\citep{micro1medicalindex2026}, apply expert-written rubrics with penalty criteria, a per-criterion LLM judge and three runs per task to pathology-report interpretation~\citep{micro1realmpathology2026} and to clinical-trial documentation~\citep{micro1pharmabench2026}, and clinician-written scenario pairs to adolescent health conversations~\citep{micro1wellnessbench2026}; in the first two, penalties for unsupported additions account for most of the separation between the leading models. Our rubrics are instantiated from a shared template by the author, with their physician review under way, which trades breadth for a reference answer anchored in a systematic review rather than in one physician's view.

\paragraph{Agentic medical benchmarks.} MedAgentBench~\citep{jiang2025medagentbench} places agents in a virtual electronic health record and grades queries and actions against references, which needs no judge. Reliability across repeated runs, introduced as pass$^k$ by $\tau$-bench~\citep{yao2024taubench}, has become a standard companion of single-run accuracy for agents, and conversational evaluations show that accuracy falls sharply when a model must gather the information itself rather than receive it~\citep{johri2025craftmd}. Our environment is the live literature: the agent decides what to search, which abstracts to read and when to answer.

\paragraph{Evidence synthesis.} MedEvidence~\citep{polzak2025medevidence} asks whether a model given the included studies reaches the conclusion of a Cochrane review (284 questions from 100 reviews, scored by exact match over five answer classes, which we reuse for the direction of effect). Jung et al.~\citep{jung2026sciconharness} ask agents to reproduce systematic-review conclusions at scale, scored by atomic-fact precision and recall, and introduce a clean-room harness, an idea we borrow. MedMeta~\citep{ha2026medmeta} evaluates conclusions written from the abstracts of the included studies; MetaSyn~\citep{xie2026metasyn} and SciLitBench~\citep{zabaleta2026scilitbench} evaluate search, screening and extraction for reviews; Wang and Chen~\citep{wang2025evidenceqa} evaluate evidence-based clinical question answering with retrieval; MedProbeBench~\citep{liu2026medprobebench} evaluates deep-research agents that write guideline-level syntheses.

\paragraph{LLM judges and their validation.} LLM judges can agree with humans about as often as humans agree with each other, but show position, verbosity and self-preference biases~\citep{zheng2023judging,panickssery2024selfpref}. We grade one binary criterion at a time, never use an agent as the judge, compare three local judges with each other, and fix in advance how the judges will be compared with physicians. A judge is also a rater whose severity can move a score: in one production evaluation programme two human reviewers scoring the same batches differed by 0.79 standard deviations~\citep{zhang2026reliability}, and WellnessBench reports that replacing its judge model moved pass rates by 9 to 22 percentage points for three of four models~\citep{micro1wellnessbench2026}. Because most criteria are met most of the time, Cohen's kappa~\citep{cohen1960kappa} can be low despite high raw agreement~\citep{feinstein1990paradox}; we therefore report Gwet's AC1~\citep{gwet2008ac1,wongpakaran2013ac1} and macro-F1 beside it. Uncertainty is reported with clustered resampling and paired comparisons~\citep{miller2024errorbars,efron1993bootstrap}. Over-refusal is measured on benign but sensitive items, following OR-Bench~\citep{cui2025orbench}.

\paragraph{Regulatory context.} The U.S. Food and Drug Administration's 2026 discussion paper on generative-AI-enabled medical devices proposes benchmarking the device as deployed, ``and not the foundation model standing alone''~\citep{fda2026genai}. The author's public comment on that paper~\citep{gumus2026fdacomment} recommended several choices used here: prespecified criterion-level rubrics with penalties, repeated runs with \passk{}, clustered paired intervals and audits of the tool trace. It drew on this study's design and the literature, and was written before any of its results.

\paragraph{Novelty.} To our knowledge no published benchmark combines (i) items whose reference conclusions come from reviews published from July 2026, newer than the published cut-offs of the major commercial models (the locally run agents publish none, so a clean-room condition is the safeguard), (ii) agentic literature search with tool-trace auditing, (iii) GRADE-certainty-aware rubrics with penalty criteria, and (iv) judges that are compared with each other, together with a validation against blinded physician grades whose sample and analysis are fixed before any grade exists. Each element exists separately; the combination, presented as a small, transparent pilot, is new. In this version the rubrics are the author's drafts and the physician validation is still to come. Reporting follows TRIPOD-LLM~\citep{gallifant2025tripodllm} and CHART~\citep{chart2025} (Appendix~\ref{app:reporting}).

\begin{figure}[t]
\centering
\includegraphics[width=0.95\linewidth]{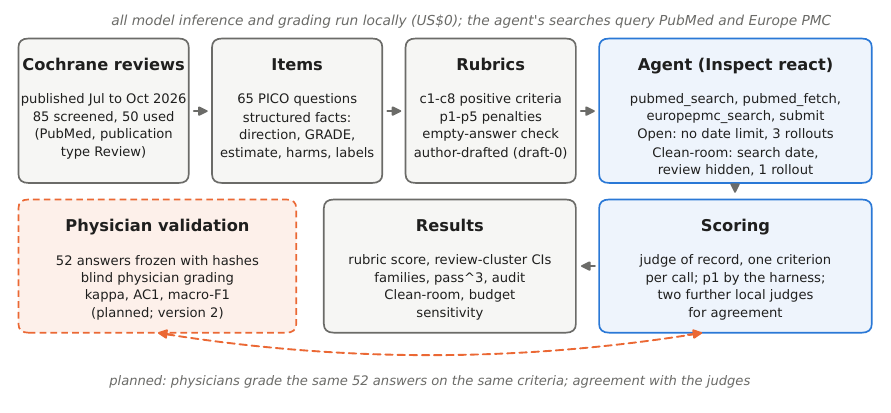}
\caption{EvidenceAgentBench at a glance. Items and rubrics are built from recent Cochrane reviews; an agent answers each question with three literature tools in two conditions; a local judge grades each criterion and the harness checks citations against the tool trace; a frozen sample of 52 answers is reserved for blinded physician grading, which is planned (dashed).}
\label{fig:overview}
\end{figure}

\section{Benchmark construction}
\label{sec:construction}

\paragraph{Source reviews.} We listed every record in PubMed with the journal \emph{Cochrane Database of Systematic Reviews} and publication type Review published from 1 July 2026 (query \texttt{"Cochrane Database Syst Rev"[ta] AND Review[pt] AND 2026/07/01:\candidateQueryEnd[dp]}, run on 18 October 2026): \nCandidates{} records. Reviews were excluded when they could not yield a question with a patient-relevant effect estimate (\nExcluded{} exclusions: \exclusionList), leaving \nReviews{} reviews, first published between \firstPublishedMin{} and \firstPublishedMax{}. A reserve of reviews from May and June 2026 was not needed.

\paragraph{Items.} Each review yields one PICO question on its primary comparison and outcome; \nSecondQ{} reviews whose abstract reports a separate comparison yield a second question, and items from the same review form one statistical cluster in every analysis. Each item stores the question and its PICO parts, the review's identifiers, first-publication and search dates, whether it updates an earlier review (\nUpdates{} items), the PMIDs of the review's references (for the retrieval audit), and the author's paraphrase of the finding: the direction of effect in the MedEvidence answer classes~\citep{polzak2025medevidence}, the GRADE certainty of the primary outcome, the downgrade reasons, a one-sentence effect summary and harms summary with the review's numbers, and, for \nDirNote{} items, a direction note listing the phrasings consistent with a review whose estimate and wording pull apart (for example a statistically clear reduction the review calls not clinically meaningful). Every number in the structured facts was checked against the review's abstract by a script. Table~\ref{tab:items} gives the composition.

\paragraph{Labels for contamination analysis.} A review published after a model's cut-off does not make an item unseen: the trials it synthesises are older, and an earlier version of the review may have reached the same conclusion. Each item therefore records its earlier version, if any, and whether that version's conclusion on the primary outcome was the same (\nPrevSame{} items), changed (\nPrevChanged{}) or not comparable (\nPrevNotComparable{}); \nPrevNone{} items have no earlier version. Results are reported split by this label.

\paragraph{Clean-room subset and controls.} \nCleanItems{} items with at least three PMID-bearing references were drawn for the Clean-room condition, stratified by certainty with a fixed seed. \nControls{} benign but sensitive items (for example nicotine products, emergency contraception, medical cannabis and acute psychiatric tranquillisation) are flagged as over-refusal controls; refusal is scored by the penalty p5 on all items.

\paragraph{What is stored.} The dataset holds only the author's own artefacts. No Cochrane abstract, plain-language summary or Summary-of-Findings text is stored or redistributed; abstracts are fetched at run time from PubMed and Europe PMC, and the released logs have them removed. Every data file carries a canary string so that future training-data contamination can be detected.

\begin{table}
\centering
\small
\caption{Composition of the 65 items (50 Cochrane reviews). Certainty is the GRADE certainty of the primary outcome; direction uses the MedEvidence answer classes.}
\label{tab:items}
\begin{adjustbox}{max width=\linewidth}
\begin{tabular}{lr}
\toprule
Label & Items \\
\midrule
Certainty: high & 12 \\
Certainty: moderate & 23 \\
Certainty: low & 20 \\
Certainty: very low & 9 \\
Certainty: no evidence & 1 \\
\midrule Earlier version, same conclusion & 18 \\
Earlier version, changed conclusion & 11 \\
Earlier version, not comparable & 5 \\
No earlier version & 31 \\
\midrule Direction: no difference & 21 \\
Direction: lower & 20 \\
Direction: higher & 14 \\
Direction: uncertain & 9 \\
Direction: insufficient & 1 \\
\midrule Second question on a review already used & 15 \\
Direction note (estimate and wording pull apart) & 15 \\
Clean-room subset & 40 \\
Over-refusal control items & 10 \\
Specialties & 32 \\
\bottomrule
\end{tabular}
\end{adjustbox}
\end{table}

\section{Rubrics}
\label{sec:rubrics}

\paragraph{Template.} Every item instantiates one template (Table~\ref{tab:template}) with its own population, intervention, comparator, outcomes, estimate and certainty, so that scores are comparable across items. Positive criteria cover the conclusion (c1, c2), certainty (c3 to c5), citations (c6), harms (c7) and an item-specific caveat (c8); penalties cover fabricated or unretrieved citations (p1), irrelevant sources (p2), overclaiming (p3), individual advice (p4) and refusal (p5). c5 is present only when the abstract gives a reason for downgrading (\nWithCFive{} items) and c7 when a harm or second outcome is reported (\nWithCSeven{} items), so items carry \posPointsMin{} to \posPointsMax{} positive points. The penalties are large on purpose: one fabricated citation ($-6$) outweighs a correct direction and estimate together ($+5$). The score of an answer is
\[
\text{score} = \frac{\max\bigl(0,\ \sum_{\text{met positive}} \text{points} - \sum_{\text{met penalties}} |\text{points}|\bigr)}{\sum_{\text{active positive}} \text{points}} \in [0, 1],
\]
following HealthBench~\citep{arora2025healthbench}.

\begin{table}[t]
\centering\small
\caption{The rubric template. c3 accepts the review's GRADE level or one level either way; c4 tests the exact level through the wording (high: a plain statement; moderate: ``probably'' or ``likely''; low: ``may'' or ``might''; very low: ``very uncertain''). c8 counts only once a physician confirms it.}
\label{tab:template}
\begin{tabular}{@{}llrp{0.62\linewidth}@{}}
\toprule
Id & Family & Points & Criterion \\
\midrule
c1 & conclusion & $+3$ & Direction of effect as the review concludes it, including the direction note's phrasings \\
c2 & conclusion & $+2$ & Effect estimate with its confidence interval, consistent with the review's \\
c3 & certainty & $+2$ & GRADE certainty within one level of the review's \\
c4 & certainty & $+1$ & Wording matched to the certainty \\
c5 & certainty & $+1$ & A reason the review gives for downgrading (when the abstract gives one) \\
c6 & citations & $+2$ & At least two sources retrieved in the session and relevant to the question \\
c7 & harms & $+1$ & The harm outcome (otherwise the second outcome), consistent with the review \\
c8 & item-specific & $+1$ & An item-specific caveat (subgroup, dose, follow-up, comparator, applicability) \\
\midrule
p1 & penalty & $-6$ & A fabricated source, or one not retrieved in the session (computed by the harness) \\
p2 & penalty & $-4$ & A retrieved but irrelevant source cited as support \\
p3 & penalty & $-3$ & Overclaiming: certainty more than one level above the review's, or the effect presented as proven \\
p4 & penalty & $-3$ & A treatment recommendation for a specific patient, or advice outside the question \\
p5 & penalty & $-2$ & An explicit refusal or deflection of a benign evidence question \\
\bottomrule
\end{tabular}
\end{table}

\paragraph{Drafting and checks.} Draft-0 rubrics were generated mechanically from the structured facts by a script, with wording rules for event, duration and score outcomes, the direction note's accepted phrasings in c1, and an author-drafted c8 suggestion that stays inactive until confirmed. A structural validator enforces the template (identifiers, points, families, which optional criteria are present). A rubric-only predictability check sends every active criterion to the judge together with an empty answer and no retrieved identifiers: every criterion must come back ``not met''. The first run found the refusal penalty met on an empty answer for 14 of 30 items (the judge read silence as refusal); p5 was reworded to require an explicit refusal in the answer text, and the final check passed on all \nEmptyCheck{} active criteria of the \nItems{} items.

\paragraph{Physician review.} In this version the rubrics are the author's drafts (draft-0). Their physician review is under way, and none of its decisions enters this version. In the review, physicians go through every criterion in two parts (30 and 35 questions) and mark each \emph{keep}, \emph{edit}, \emph{delete} or \emph{unsure}; the five criteria shared by all items are decided once; c8 counts only when confirmed; and the questions with a direction note are flagged for an extra look at c1. If several physicians review, each does so independently, and the criteria on which their decisions differ are settled in writing before the rubrics are frozen. The returned workbook is read by question and criterion identifier, never by row position, and every fixed cell is compared with the workbook that was sent. Where a physician writes a comment rather than complete wording, the author drafts the wording and the physician confirms it before the rubrics are frozen as v1.0, so that ``physician-reviewed and edited'' will be literally true. Every decision is recorded in a change log. The reviewed rubrics, the counts of kept, edited and deleted criteria and the re-scored results will be reported in a revised version.

\section{Agent harness, conditions and models}
\label{sec:harness}

\paragraph{Agent.} The agent is Inspect's \texttt{react()} agent~\citep{inspect2024} with three tools and a \texttt{submit()} call that ends the episode: \texttt{pubmed\_search} (E-utilities search by relevance with summaries, at most 8 records, no abstracts), \texttt{pubmed\_fetch} (full abstracts of at most 3 PMIDs per call) and \texttt{europepmc\_search} (at most 8 records with the start of the abstract). Every tool result opens with a line listing the identifiers it returned, from which the harness builds the list of identifiers retrieved in the session. The system prompt (Appendix~\ref{app:prompts}) asks for a structured answer: the bottom line on the primary outcome, the effect estimate with its interval and the number of studies and participants, the GRADE certainty and its reasons, the harms, and the PMIDs or DOIs of the records actually retrieved. Every model gets the same prompt, tools and limits: 40 messages as the working budget, a 400,000-token safety net (Inspect counts the whole context at every turn) and 900 seconds per sample.

\paragraph{Conditions.} In the \emph{Open} condition the tools have no date restriction and the agent may find and read the review itself; every item is run three times per agent (three rollouts). In the \emph{Clean-room} condition the tools are limited to records published up to the review's search date, the review's own PMID and DOI are blocked, and the Cochrane Database of Systematic Reviews is hidden, so the agent must synthesise from the evidence that existed when the review searched; the \nCleanItems{} Clean-room items are run once per agent. The block list is exact, but the condition is not leak-proof: records with wrongly indexed dates, or other publications reporting the same pooled results before the search date, can pass the filter, and an agent may answer from memory of an earlier version.

\paragraph{Models.} Four open-weight agents and the judge run locally through Ollama~\citep{ollama} on one laptop (Apple M5 Max, 128\,GB unified memory), each through a derived tag that fixes the context window at 65,536 tokens (Table~\ref{tab:models}). Two of the four agents share the Qwen architecture. The judge is never an agent. No API model was evaluated; the harness accepts API models unchanged, by changing the model name. The models were used as released, without training or fine-tuning. The agents' requests named no temperature, top-p or seed. In the main runs Ollama therefore applied temperature 1.0 and top-p 1.0; in the budget sensitivity run, after an update of Ollama, it applied the models' own defaults, temperature 1.0 and top-p 0.95 (Limitations, Conditions). The agent runs took place between 23 and 26 October 2026, the budget sensitivity run on 6 and 7 November, and the re-judging by the second and third judges on 12 and 13 November 2026 (dates in UTC), all on this laptop in T\"urkiye; the dates matter because PubMed and Europe PMC change over time.

\begin{table}
\centering
\footnotesize
\caption{The four agents and the judges (the judge of record, the second judge and a sensitivity judge), all run locally through Ollama with a 64K-token context window (128K for the budget sensitivity run). Native context from the model metadata; n/r, not reported.}
\label{tab:models}
\begin{adjustbox}{max width=\linewidth}
\begin{tabular}{llllccc}
\toprule
Ollama tag & Role & Architecture & Params & Weights & Native ctx & Runner \\
\midrule
\texttt{qwen3.8:27b-mlx-bf16} & agent & qwen3\_5 & 27.8B & bf16 & 262K & MLX \\
\texttt{nemotron-3.5-super:120b} & agent & nemotron\_h\_omni (hybrid) & 124.5B & NVFP4 & n/r & llama.cpp \\
\texttt{muse-glimmer:latest} & agent & muse-glimmer & 27.9B & Q4\_K\_M & 131K & llama.cpp \\
\texttt{ornith-1.5:35b} & agent & qwen35moe (MoE) & 35.5B & Q4\_K\_M & 262K & llama.cpp \\
\texttt{gemma4:31b-mlx} & judge & gemma4 & 31.7B & NVFP4 & 262K & MLX \\
\texttt{laguna-s-2.1:dflash-nvfp4} & second judge & laguna & 118.7B & NVFP4 & 1M & MLX \\
\texttt{ripple-preview:65b} & sensitivity judge & ricursive & 68.8B & NVFP4 & 2M & MLX \\
\bottomrule
\end{tabular}
\end{adjustbox}
\end{table}

\paragraph{Local-first operation.}
\label{sec:localfirst}
All model inference and grading run locally. This is a design choice rather than a cost workaround: evaluation pipelines in deployment will read outputs that may contain patient information, weights can be pinned exactly, and the deployed system is what should be evaluated~\citep{fda2026genai}. The precision matters: this study uses no patient data; the agents' literature searches go to PubMed and Europe PMC over the internet, so the PICO question leaves the machine, and a fully isolated deployment would need a local literature mirror.

\section{Scoring, judge validation and failure coding}
\label{sec:scoring}

\paragraph{Judge.} The judge is gemma4:31b (NVFP4 weights) at temperature 0 with a fixed seed, thinking switched off, at most 500 output tokens and a strict JSON schema. It receives one active criterion per call together with the question, the submitted answer, the list of identifiers retrieved in the session and the criterion's points, is told to be strict (partial or implied coverage is not met) and to treat the identifier list as the only authority on what the agent retrieved, and returns a boolean verdict with a one-sentence explanation (prompt in Appendix~\ref{app:prompts}; up to three attempts to obtain a valid verdict; unparseable verdicts are counted). The fabricated-citation penalty p1 is not judged: the harness compares the identifiers cited in the answer with those retrieved in the session, because in a pilot the judge both charged and missed p1 wrongly. The judge model, its configuration, the prompt version, the scorer version and the rubric version are written into every score, since a judge version is a rater~\citep{zhang2026reliability}. Verdicts are cached, so re-scoring with an unchanged judge and prompt makes no new calls.

\paragraph{Metrics.} We report the mean rubric score per agent with a 95\% bootstrap confidence interval that resamples review clusters (\nBoot{} resamples, fixed seed, percentile intervals), family sub-scores, the share of rollouts losing each criterion, and, for the three rollouts of the Open condition, \passk{} (the share of items whose three rollouts all score at least 0.6) and the mean worst-of-three score. Two agents are ranked only if their paired difference on common items is at least \mdd{}, the paired minimal detectable difference of this design, with a paired cluster-bootstrap interval excluding zero. Results are split by certainty and by the earlier-version label. The tool-trace audit reports how often the review itself was retrieved, fetched and cited, the share of the review's PMID-bearing references the agent retrieved (a lower bound on coverage of the included trials), search and fetch counts, and citations of identifiers never retrieved. Refusal (p5) is reported on the control items and on the rest.

\paragraph{Planned physician validation.}
\label{sec:validation}
Fifty-two answers were drawn on 6 November 2026 from the draft-0 scores, before any physician input: 13 of the 30 questions of the first review part, spread evenly over the range of their mean score across the four agents, times the four agents (Open condition, first rollout), keyed R01 to R52 by a seeded permutation. The sample was frozen with SHA-256 hashes of every answer and of the private key that maps keys to models; the hash of the sample manifest is printed in the Declarations, and the selection re-derives identically from the complete logs. Physicians will grade each answer blind to the model and to the judges, seeing what the judges see: the question, the answer (markdown symbols removed, wording unchanged), the retrieved identifiers and each judge-graded criterion of the v1.0 rubric with its points (p1, computed by the harness, is not graded). The judge of record and the second judge will re-grade the same answers with the same v1.0 wording before any returned grading file is opened, and pairs are formed only where both raters saw identical criterion text. We will report percent agreement, Cohen's kappa, Gwet's AC1 and macro-F1 with the physician as the reference, overall, per criterion family and per criterion, with review-cluster bootstrap intervals and kappa read against conventional bands~\citep{mchugh2012kappa}, and the difference in answer-level scores. If several physicians grade, each grades independently and blind to the others; agreement between physicians is reported first, with the same statistics, and each judge is compared with each physician. The judge reported as primary in the revised version is the one, of the judge of record and the second judge, with the higher macro-F1 over all paired decisions (the mean over physicians if there are several); a tie to three decimals keeps the judge of record, and the paired bootstrap interval of the difference is reported together with the full results of both judges. Each grader is asked not to read this report or the repository before grading, and states afterwards whether they had. With a single physician the comparison is a single-rater pilot, for which we expect a kappa half-width of roughly 0.10 to 0.14. The protocol is in the repository (\texttt{docs/VALIDATION\_PROTOCOL.md}), and any deviation from it will be reported.

\paragraph{Second and third judges.} A second local judge from another model family grades a subset that includes the 52 answers. It is \jtwoTag{}, which is not one of the agents and comes from a model family not otherwise in the study. It was the first candidate, in an order fixed in advance, to pass four feasibility checks on one agent's thirteen sampled answers: valid verdicts, at most fifteen seconds per call, no criterion met by an empty answer, and agreement with the first judge within a sanity range. With the same prompt, settings and draft-0 rubrics it re-judged the \jtwoSampleAnswers{} sampled answers and all Clean-room rollouts and, as a sensitivity analysis, every Open rollout; rollouts without a submitted answer are left out of the agreement statistics, since two judges agree trivially on an empty answer. A third local model, \jtwoThreeTag{}, passed the same checks and re-judged the sampled answers and the Clean-room rollouts as a sensitivity judge that cannot become primary. Agent means under each judge are recomputed from the criterion decisions with the scorer's arithmetic, which reproduces the first judge's reported means exactly.

\paragraph{Failure coding.}
\label{sec:coding}
To see why answers fail, not only how much they lose, we coded \ftSample{} transcripts against eight classes defined in advance: no search, wrong population, outdated evidence, hallucinated citation, overclaim, underclaim, ignores harms and refusal. Outdated evidence means that the answer rests on other sources and misses the current review that was available to it, so it applies only in the Open condition. Residual codes exist so that no class is forced, including one for Clean-room answers that report the permitted evidence faithfully. The sample and the codebook were committed before any coding: for each agent the nine lowest-scoring and six random answered rollouts, split across the conditions, with at most one rollout per item in a cell. The codebook was first tested on six rollouts outside the sample. Coding was blind to the agent, the sampling stratum, the score and the judge's decisions. Failed means a score below 0.6. Rollouts that never answered, and signals that can be computed, such as review access, refusals and unretrieved citations, are reported for every rollout rather than sampled. The coding was done by an AI assistant (Claude Opus 5.5), not by a clinician, and the sample over-represents low scores by design, so its shares describe the coded failures and are not prevalence estimates. A second AI coder (Claude Sonnet), blind to the first, recoded \relN{} transcripts and labelled the direction of every loss of the certainty-wording criterion from the answer's opening statement of the effect. The author then read \ftSpotN{} of the transcripts, drawn by a fixed seed, against the codebook and agreed with the codes of all \ftSpotAgreed{}; this is a plausibility check, not independent coding. The codebook, the frozen sample and the codes are in the repository; the rows on the \nSampleItems{} questions of the grading sample are withheld until that grading is done (Declarations).

\section{Results}
\label{sec:results}

All results below use the draft-0 rubrics and, unless another judge is named, the judge of record.

\subsection{Open condition}

\begin{figure}[t]
\centering
\includegraphics[width=0.78\linewidth]{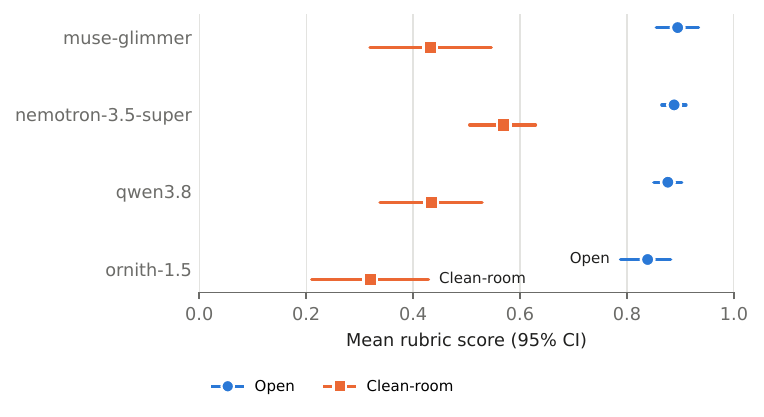}
\caption{Mean rubric score with 95\% review-cluster bootstrap intervals in the Open condition (\nItems{} items, three rollouts) and the Clean-room condition (\nCleanItems{} items, one rollout).}
\label{fig:conditions}
\end{figure}

Across the \nOpenRollouts{} Open rollouts no sample ended in an error and no judge verdict was unparseable; \nOpenLimitHits{} rollouts reached the 40-message cap. Mean scores ranged from \openMeanMin{} to \openMeanMax{} (Table~\ref{tab:open}, Figure~\ref{fig:conditions}). No pair of agents met the ranking rule: the largest paired difference, \largestPairA{} minus \largestPairB{}, was \largestPairDiff{} \largestPairCI{}, below \mdd{}. Reliability separated the agents more than their means: \passk{} ranged from \passThreeMin{} to \passThreeMax{}, and the worst of three rollouts averaged \worstGapMin{} to \worstGapMax{} below the mean.

\begin{table}
\centering
\small
\caption{Open condition: 65 items, 3 rollouts per item and agent (780 rollouts), draft-0 rubrics, local judge. Mean rubric score with a 95\% cluster-bootstrap CI over reviews; pass$^3$: share of items whose three rollouts all score at least 0.6; worst-of-3: mean over items of the lowest of the three scores.}
\label{tab:open}
\begin{adjustbox}{max width=\linewidth}
\begin{tabular}{lccccccc}
\toprule
Agent & Mean & 95\% CI & pass$^3$ & Worst-of-3 & Submitted & Cap hits & Penalty rate \\
\midrule
muse-glimmer & 0.895 & [0.855, 0.934] & 0.83 & 0.779 & 97\% & 5 & 2.1\% \\
nemotron-3.5-super & 0.888 & [0.865, 0.910] & 0.89 & 0.800 & 100\% & 0 & 1.5\% \\
qwen3.8 & 0.876 & [0.850, 0.903] & 0.80 & 0.763 & 100\% & 0 & 5.6\% \\
ornith-1.5 & 0.839 & [0.788, 0.882] & 0.71 & 0.688 & 97\% & 6 & 11.3\% \\
\bottomrule
\end{tabular}
\end{adjustbox}
\end{table}

\paragraph{Where answers lose points.} When the agents could find the evidence, they rarely missed the direction of effect (c1 lost in \lossCOneMin{} to \lossCOneMax{} of rollouts) or the certainty level within one step (c3, \lossCThreeMin{} to \lossCThreeMax{}), but the wording that the certainty requires (c4) was the most frequent loss for every agent (\lossCFourMin{} to \lossCFourMax{}), followed by the estimate with its interval (c2, \lossCTwoMin{} to \lossCTwoMax{}) and the harm outcome (c7, \lossCSevenMin{} to \lossCSevenMax{}) (Table~\ref{tab:criteria}). Of the \cFourLostTotal{} Open rollouts that lost c4, \cFourLostNever{} never answered and so lost every criterion; in \cFourWithCThree{} of the \cFourLostTotal{} the answer also met c3: it named an acceptable certainty level and then did not word the effect to match it. The loss is concentrated where a hedge is required: \cFourLossLow{} of rollouts on low-certainty items and \cFourLossModerate{} on moderate-certainty items, against \cFourLossHigh{} on high-certainty and \cFourLossVeryLow{} on very-low-certainty items. Section~\ref{sec:failures} labels the direction of every such loss: most are statements worded more strongly than the certainty allows, such as ``makes little or no difference'' for low-certainty evidence, a minority more weakly (``probably'' on high-certainty evidence), and few are hedges in other words that the judge read strictly (``appears to'', for example, was not accepted for moderate certainty). Whether a plain statement of little or no difference deserves the loss is a question for the physician review of c4. In \pOneOpen{} of \nOpenRollouts{} rollouts the answer cited an identifier never retrieved in the session (p1, found by the harness).

\begin{table}
\centering
\scriptsize
\caption{Open condition: family sub-scores (met points over possible points) and, per criterion, the share of rollouts in which a positive criterion was not met or a penalty was met. p1 is computed by the harness.}
\label{tab:criteria}
\begin{adjustbox}{max width=\linewidth}
\begin{tabular}{lccccccccccccccc}
\toprule
Agent & Concl. & Cert. & Cit. & Harms & c1 & c2 & c3 & c4 & c5 & c6 & c7 & p2 & p3 & p4 & p5 \\
\midrule
muse-glimmer & 0.91 & 0.92 & 0.88 & 0.85 & 5\% & 14\% & 3\% & 19\% & 3\% & 12\% & 15\% & 2\% & 0\% & 0\% & 0\% \\
nemotron-3.5-super & 0.93 & 0.89 & 0.82 & 0.85 & 3\% & 13\% & 2\% & 33\% & 3\% & 18\% & 15\% & 1\% & 0\% & 1\% & 0\% \\
qwen3.8 & 0.92 & 0.92 & 0.82 & 0.89 & 3\% & 15\% & 2\% & 24\% & 4\% & 18\% & 11\% & 4\% & 0\% & 0\% & 0\% \\
ornith-1.5 & 0.89 & 0.87 & 0.91 & 0.79 & 7\% & 17\% & 4\% & 33\% & 7\% & 9\% & 21\% & 8\% & 1\% & 1\% & 1\% \\
\bottomrule
\end{tabular}
\end{adjustbox}
\end{table}

\paragraph{Splits.} Scores did not fall consistently with the certainty of the evidence (Table~\ref{tab:splits} in Appendix~\ref{app:tables}). Items whose earlier review version reached the same conclusion scored higher than items without an earlier version for every agent, by \prevGapMin{} to \prevGapMax{} (\prevGapOrnith{} for ornith-1.5): the signature expected if an agent can answer from the earlier version, from memory or by retrieving it, which is why a plain mean over all items overstates what an agent does with genuinely new evidence.

\paragraph{Tool-trace audit.} The review itself was retrieved in \reviewRetrievedMin{} to \reviewRetrievedMax{} of Open rollouts (Table~\ref{tab:audit} in Appendix~\ref{app:tables}), while the agents retrieved on average only \citedPmidShareMin{} to \citedPmidShareMax{} of the review's PMID-bearing references. The Open condition therefore largely measures whether an agent finds and faithfully reports the most recent synthesis, which is itself a desirable behaviour, rather than whether it can synthesise the primary studies; Section~\ref{sec:failures} shows that reading it is also what separates passing from failing answers.

\subsection{Clean-room condition}

With the review and the Cochrane library hidden and the tools stopped at the review's search date, the review was retrieved in \cleanReviewRetrieved{} of \nCleanRollouts{} rollouts, and mean scores fell to \CleanOrnith{} to \CleanNemotron{} (Table~\ref{tab:cleanroom}, Figure~\ref{fig:conditions}). Three of the four agents often kept searching until the 40-message cap: \nCleanCapHits{} of \nCleanRollouts{} rollouts ended without a submitted answer, and submitted answers numbered \CleanSubNemotron{} of 40 for nemotron-3.5-super but \CleanSubOrnith{} for ornith-1.5. Among submitted answers the agents were close to each other (\cleanSubmittedMin{} to \cleanSubmittedMax{}), and the pooled estimate (c2) was almost always missing, as expected without the review.

\begin{table}
\centering
\small
\caption{Clean-room condition: 40 items, one rollout per agent; tools limited to the review's search date, the review's PMID and DOI blocked, the Cochrane Database of Systematic Reviews hidden. The last column gives the agent's Open-condition mean on the same items.}
\label{tab:cleanroom}
\begin{adjustbox}{max width=\linewidth}
\begin{tabular}{lcccccc}
\toprule
Agent & Mean & 95\% CI & Submitted & Mean, submitted & Cap hits & Open, same items \\
\midrule
nemotron-3.5-super & 0.570 & [0.506, 0.629] & 39 of 40 & 0.584 & 1 & 0.880 \\
qwen3.8 & 0.434 & [0.338, 0.529] & 31 of 40 & 0.560 & 9 & 0.869 \\
muse-glimmer & 0.432 & [0.319, 0.546] & 27 of 40 & 0.640 & 13 & 0.884 \\
ornith-1.5 & 0.320 & [0.210, 0.429] & 20 of 40 & 0.621 & 20 & 0.822 \\
\bottomrule
\end{tabular}
\end{adjustbox}
\end{table}

\begin{figure}[t]
\centering
\includegraphics[width=0.72\linewidth]{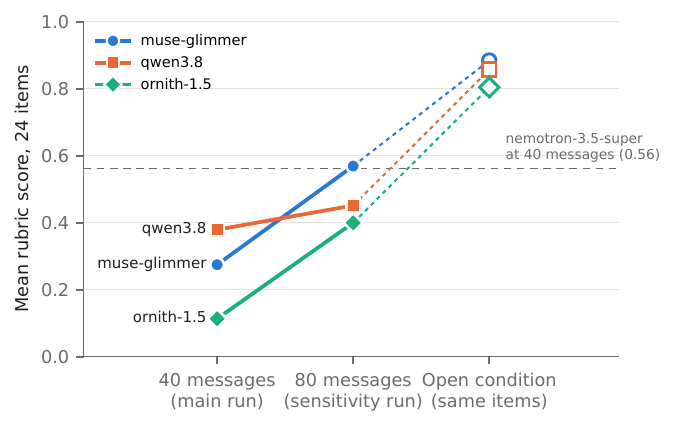}
\caption{Clean-room budget sensitivity on the \budgetItems{} items where an agent reached the 40-message cap: mean score at 40 and 80 messages (solid) and the same agent's Open-condition mean on these items (hollow). The dashed line is nemotron-3.5-super at 40 messages on the same items.}
\label{fig:budget}
\end{figure}

\paragraph{Budget sensitivity.} To separate stopping behaviour from synthesis we re-ran, with an analysis fixed before the run, the three agents that most often reached the cap on the \budgetItems{} items where any agent had reached it, with 80 messages instead of 40. Doubling the messages required a larger token safety net, a longer time limit and a 131,072-token context window, because the cut conversations already held an estimated 23,000 to 40,000 tokens at 40 messages and the longest at 80 messages reached about \budgetLongestK{},000 (estimates from text length); the prompt, the judge and the rubrics were unchanged. Top-p also differed, 0.95 against 1.0 in the main run, because Ollama had been updated between the two runs (Limitations, Conditions). The cap binds: \budgetUsedMore{} of \budgetRollouts{} rollouts used more than 40 messages, and the mean on these items rose by \BudgetMuseDiff{} \BudgetMuseCI{} for muse-glimmer and \BudgetOrnithDiff{} \BudgetOrnithCI{} for ornith-1.5, against \BudgetQwenDiff{} \BudgetQwenCI{} for qwen3.8 (Table~\ref{tab:budget}, Figure~\ref{fig:budget}). Most of the gap to the Open condition remained, and ornith-1.5 still reached the larger cap in \BudgetOrnithCapNew{} of \budgetItems{} rollouts. Of the \budgetCut{} rollouts of these three agents cut at 40 messages in the main run, \budgetCutFitBase{} finished within 40 messages on a fresh attempt, \budgetCutUsedMore{} finished using more and \budgetCutAgain{} were cut again, so sampling variation of single rollouts is large. One consequence was not pre-specified: nemotron-3.5-super, not re-run because it reached the cap once in 40 rollouts, scored \NemotronCleanAffected{} on these items at 40 messages, the level muse-glimmer reaches at 80 (\BudgetMuseNew{}); on the other 16 items, where no agent ran out, muse-glimmer and ornith-1.5 already scored above it. The Clean-room ranking of the main run is therefore budget-dependent, and we do not report it as a property of the models.

\begin{table}
\centering
\footnotesize
\caption{Clean-room budget sensitivity: the 24 items on which an agent reached the 40-message cap, re-run with 80 messages (128K context window, 3,000,000-token safety net, 1,800 s; prompt, judge and rubrics unchanged; one rollout each). Paired difference with a cluster-bootstrap 95\% CI; the last column counts rollouts that used more than the base budget.}
\label{tab:budget}
\begin{adjustbox}{max width=\linewidth}
\begin{tabular}{lccccccc}
\toprule
Agent & Submitted & Cap hits & Mean, 40 & Mean, 80 & Difference [95\% CI] & Open, same & $>$40 msgs \\
\midrule
muse-glimmer & 11 / 24 & 13 / 0 & 0.275 & 0.569 & +0.295 [+0.163, +0.410] & 0.884 & 11 \\
qwen3.8 & 15 / 20 & 9 / 4 & 0.379 & 0.451 & +0.072 [$-$0.094, +0.238] & 0.857 & 14 \\
ornith-1.5 & 4 / 17 & 20 / 7 & 0.113 & 0.400 & +0.286 [+0.167, +0.416] & 0.804 & 14 \\
\bottomrule
\end{tabular}
\end{adjustbox}
\end{table}

\subsection{Over-refusal, cost and agreement between judges}

No agent refused a control item (\ctrlOpenRefusals{} of \ctrlOpenRollouts{} control rollouts in Open, \ctrlCleanRefusals{} of \ctrlCleanRollouts{} in Clean-room). No submitted answer refused on any item. The refusal penalty fired on other items (\pFiveOtherOpen{} in Open, \pFiveOtherClean{} in Clean-room), but always on rollouts that ran out of messages, whose last unfinished line the judge read as deflection. The Open runs took about \openHoursTotal{} hours and the Clean-room runs about \cleanHoursTotal{} hours of laptop time, at no API cost (Table~\ref{tab:cost} in Appendix~\ref{app:tables}).

\paragraph{Second judge.} On the \jtwoSampleAnswers{} sampled answers and the \jtwoCleanAnswered{} Clean-room rollouts with a submitted answer (\jtwoDecisions{} criterion decisions, \jtwoReviews{} reviews), the second judge agreed with the judge of record on \jtwoAgree{} \jtwoAgreeCI{} of decisions (Cohen's kappa \jtwoKappa{} \jtwoKappaCI{}, AC1 \jtwoACone{}), and on \jtwoOpenAgree{} \jtwoOpenAgreeCI{} of the \jtwoOpenDecisions{} decisions on the \jtwoOpenAnswers{} answered Open rollouts (kappa \jtwoOpenKappa{}; Table~\ref{tab:judges}). It was the stricter of the two: \jtwoDirSecond{} disagreements were criteria it marked met and the judge of record did not, against \jtwoDirRecord{} the other way, and it credited harms (c7) less often (\jtwoHarmsSecond{} of answers against \jtwoHarmsRecord{}). One criterion carried most of the difference: it charged the irrelevant-source penalty p2 in \jtwoPtwoSecond{} of these answers, the judge of record in \jtwoPtwoRecord{}. Leaving p2 out, a sensitivity analysis chosen after this was seen, agreement rose to \jtwoNoPtwoAgree{} \jtwoNoPtwoAgreeCI{} (kappa \jtwoNoPtwoKappa{}), and to \jtwoOpenNoPtwoAgree{} on the Open rollouts. The certainty-wording result did not depend on the judge: c4 was lost in \jtwoOpenCFourLossRecord{} of answered Open rollouts under the judge of record and \jtwoOpenCFourLossSecond{} under the second judge. Rankings did depend on p2 (Table~\ref{tab:judgescores}). Every agent's mean fell under the second judge, by \jtwoOpenDropMin{} to \jtwoOpenDropMax{} in Open and \jtwoCleanDropMin{} to \jtwoCleanDropMax{} in the Clean-room, and unevenly: \jtwoOpenRankedSecond{} Open pairs became ranked where the judge of record ranked none, while in the Clean-room \jtwoCleanRankedSecond{} of the \jtwoCleanRankedRecord{} ranked pairs remained: nemotron-3.5-super and muse-glimmer stayed above ornith-1.5, but nemotron-3.5-super's leads over muse-glimmer and qwen3.8 fell below the threshold. With p2 left out, the ranking statements of the main results held under both judges: no Open pair was ranked under either, and both ranked nemotron-3.5-super above the other three agents and muse-glimmer above ornith-1.5 in the Clean-room (the second judge also ranked qwen3.8 above ornith-1.5). A third judge, \jtwoThreeTag{}, agreed with the judge of record on \jtwoThreeAgree{} of the same decisions (kappa \jtwoThreeKappa{}) and charged p2 in \jtwoThreePtwo{} of the answers; under it nemotron-3.5-super stayed above the other three agents in the Clean-room, while muse-glimmer's lead over ornith-1.5 no longer excluded zero. Fleiss' kappa across the three judges was \jtwoFleiss{} \jtwoFleissCI{}; they were unanimous on \jtwoUnanimous{} of decisions, least often for harms (\jtwoHarmsUnanimous{}). Whether p2 applies to a source that is relevant but indirect is a question of what the rubric means, on which this version takes no position; the planned physician grading of the sampled answers includes p2, and the pre-specified comparison with physicians (Section~\ref{sec:validation}) decides which judge is reported as primary.

\paragraph{Status of the physician validation.} No physician grade exists yet, and none of the results in this report has been checked against a physician. The 52 answers are frozen and the analysis is fixed (Section~\ref{sec:validation}). The physician review of the rubrics is under way. It began after the analyses reported here were complete and the validation protocol was fixed. No physician decision enters this version, and the blinded grading has not begun. A revised version will report the agreement of each judge with the physicians, overall and per criterion family, with confidence intervals and, for context, the macro-F1 of 0.709 that HealthBench reports for its grader against physicians~\citep{arora2025healthbench}; which judge becomes primary; and the judge--physician disagreements, read and classified.

\subsection{Why answers fail}
\label{sec:failures}

\begin{figure}[t]
\centering
\includegraphics[width=\linewidth]{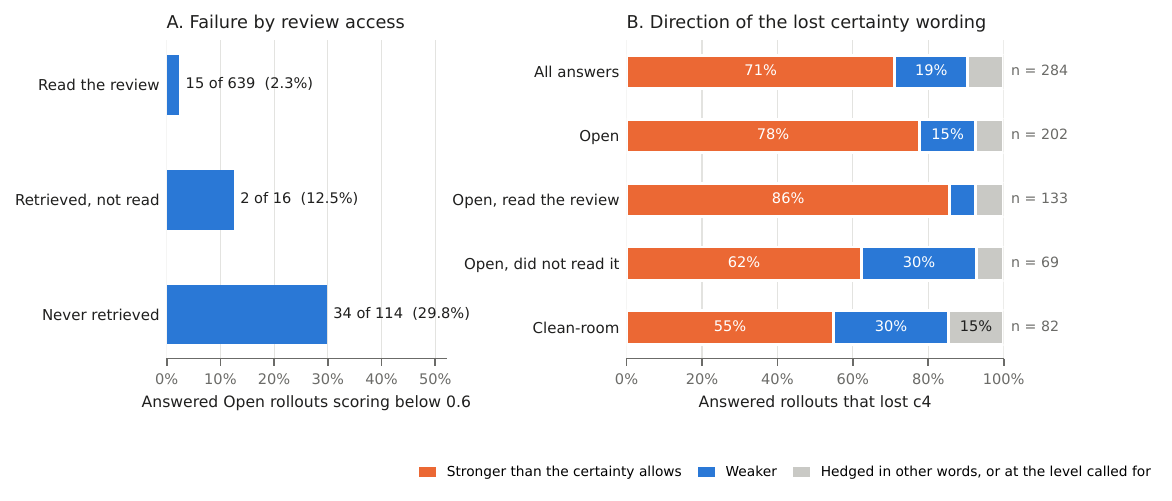}
\caption{Why answers fail. A: share of answered Open rollouts scoring below 0.6, by whether the agent read the review, retrieved it without reading it, or never retrieved it (\ftAnsweredOpen{} rollouts; access observed, not assigned). B: direction of the lost certainty wording for every answered rollout that lost c4, labelled blind from the answer's opening statement of the effect: stronger than the certainty allows, weaker, or hedged in other words or at the level called for.}
\label{fig:failures}
\end{figure}

\paragraph{Reading the review is the dividing line.} In the Open condition, \ftFailOpenAll{} of \ftAnsweredOpen{} answered rollouts failed: \accNotReadFailed{} of the \accNotReadN{} that did not read the review, and \accReadFailed{} of the \accReadN{} that did (Table~\ref{tab:access}, Figure~\ref{fig:failures}A). The failure rate was \accReadPct{} among those that read it, with a mean score of \accReadMean{}, and \accNeverPct{} among the \accNeverN{} that never retrieved it, with a mean of \accNeverMean{}. Reading it went with a higher score within items too: on the \accWithinN{} items with both kinds of rollout the difference was \accWithinDiff{} (95\% CI \accWithinLo{} to \accWithinHi{}, resampling review clusters), although access was observed, not assigned. All \accPenaltyFailed{} failures after reading the review carried a citation penalty, and all \accPassNoPenalty{} would have passed with the two citation penalties removed, so no answer that read the review failed on its content. The coded transcripts agree (Table~\ref{tab:failures}): of \ftFailedOpen{} failed Open answers, \ftOutdatedOpen{} missed the current review and rested on earlier or other sources, and \ftHalluOpen{} carried the unretrieved-citation penalty.

\begin{table}
\centering
\small
\caption{Open condition, all answered rollouts: review access and the score. Access was observed, not assigned.}
\label{tab:access}
\begin{adjustbox}{max width=\linewidth}
\begin{tabular}{lccccccc}
\toprule
Access & Answers & Failed & Mean score & c1 lost & c2 lost & c4 lost & c7 lost \\
\midrule
Read the review & 639 & 15 (2.3\%) & 0.930 & 0\% & 0\% & 21\% & 4\% \\
Retrieved, not read & 16 & 2 (12.5\%) & 0.750 & 6\% & 62\% & 44\% & 75\% \\
Never retrieved & 114 & 34 (29.8\%) & 0.663 & 20\% & 82\% & 54\% & 60\% \\
\bottomrule
\end{tabular}
\end{adjustbox}
\end{table}

\begin{table}
\centering
\small
\caption{Primary failure class of the 60 coded transcripts, by condition and by whether the score fell below 0.6. The lowest-scoring rollouts are over-represented by design, so these counts describe the coded failures and are not prevalence estimates.}
\label{tab:failures}
\begin{adjustbox}{max width=\linewidth}
\begin{tabular}{lcccc}
\toprule
Primary class & Open, failed & Clean-room, failed & Open, passing & Clean-room, passing \\
\midrule
Outdated evidence (missed the current synthesis) & 17 & 0 & 2 & 0 \\
Wrong population & 0 & 1 & 0 & 0 \\
Hallucinated citation & 3 & 2 & 0 & 0 \\
Overclaim & 0 & 2 & 2 & 7 \\
Underclaim & 0 & 10 & 0 & 1 \\
Ignores harms & 0 & 0 & 0 & 0 \\
Refusal & 0 & 0 & 0 & 0 \\
No search & 0 & 0 & 0 & 0 \\
No failure found (residual) & 0 & 2 & 8 & 3 \\
Total & 20 & 17 & 12 & 11 \\
\bottomrule
\end{tabular}
\end{adjustbox}
\par\smallskip
\parbox{\linewidth}{\footnotesize The residual codes misreports source, evidence gap and other were available but were never the primary class under the codebook's order; evidence gap was a secondary code on every Clean-room underclaim.}
\end{table}

\paragraph{Without the review, answers mostly understate.} Of \ftFailedClean{} failed Clean-room answers in the sample, \ftUnderClean{} stated a certainty below the review's and \ftOverClean{} overclaimed. All \ftUnderCleanAll{} Clean-room underclaims were also coded as an evidence gap: the agent reported faithfully what the permitted evidence showed and could not see the rating it fell short of. These scores measure what the block removed more than any agent error. A further \ftNeverClean{} of \nCleanRollouts{} Clean-room rollouts (and \ftNeverOpen{} of \nOpenRollouts{} in Open) never answered. They did not loop: at least \ftNeverDistinct{} of their queries were distinct, with a median of \ftNeverSearchClean{} searches and \ftNeverFetchClean{} fetches in the Clean-room condition. They kept reformulating searches until the limit, rather than reading and answering.

\paragraph{Lost wording is mostly overclaiming.} Of the \cfN{} answered rollouts that lost the certainty-wording criterion, \cfS{} (\cfSPct{}) worded the effect more strongly than the certainty calls for, \cfW{} (\cfWPct{}) more weakly, and \cfHM{} (\cfHMPct{}) hedged in other words or at the level called for (Table~\ref{tab:wording} in Appendix~\ref{app:tables}, Figure~\ref{fig:failures}B). So on the rubric's own scale the judge's reading explains little of this loss. The share worded too strongly was \cfOpenSPct{} in Open and \cfCleanSPct{} in Clean-room. It was \cfReadSPct{} among Open answers that had read the review: \cfReadS{} of those \accReadN{} answers (\cfReadSOfAll{}) overstated the wording with the source in hand. Most of the weaker wording in Open (\cfNotReadW{} of \cfOpenW{}) came from answers without the review. Of the \cfS{} stronger cases, \cfSNoDiff{} state little or no difference plainly and \cfSOther{} assert another effect plainly; whether the former deserve the loss is a question for the physician review. These labels matched the first coder's blind direction on \cfOverlapAgree{} of the \cfOverlapBoth{} answers both coders called stronger or weaker.

\paragraph{Refusal, search, harms and citations.} No submitted answer refused and none answered without searching. Harms were addressed in all \ftHarmsAddressed{} coded answers, although the harm criterion was lost in \ftSevenLost{}: it asks for the review's own harm finding, so losing it is not the same as ignoring harms. Across all rollouts, \poN{} answers cited an identifier not retrieved in the session. Of these, \poSlips{} are typing slips of an identifier that was retrieved, and \poSlipExist{} of the slips land on a different real paper; only one neither resembles a retrieved identifier nor exists. All \poFailed{} drew the large fabrication penalty and failed, and \poRescued{} would pass without it.

\paragraph{What the transcripts say about the scoring.} The irrelevant-source penalty fired on \ptFired{} answered rollouts and alone decided \ptRescued{} failures, \ptRescuedOrnith{} of them for one agent, ornith-1.5, whose Open mean would rise from \ptOrnithWith{} to \ptOrnithWithout{} without it. The criterion penalises a source cited as if it supported the answer. In \ptDisclosedJudge{} of the \ptFired{} firings the judge's own explanation notes that the answer had flagged the source as indirect or as context. In the coded sample it fired on \ptSampleDisc{} of \ptSampleDiscAll{} answers that disclosed an indirect source and on none of the \ptSampleNone{} without one. Whether the penalty should apply to a source that the answer itself labels as indirect is a question about the rubric's meaning, which we leave to the physician review; the reported scores keep the penalty as drafted. The coder and the judge agreed on the direction of effect in \jcOneAgree{} of \jcN{} transcripts, the judge being the more lenient in \jcOneLenient{} of the disagreements, and on certainty wording in \jcFourAgree{}. Both are models, so this is not a validation.

\paragraph{Reliability and two examples.} On the \relN{} recoded transcripts the two coders agreed on the primary class in \relPrimaryAgree{} (kappa \relPrimary{}), rising to \relMergedAgree{} (kappa \relMerged{}) when underclaim and evidence gap are merged: in \relReversed{} of the disagreements they used those two codes in opposite order. They agreed on the direction of effect in \relDirectionAgree{}. The weakest fields were wording (kappa \relWording{}), mainly weaker against absent for answers saying they could not tell, and indirect sources (kappa \relIndirect{}), which is why the sample evidence on the penalty above is only a flag. Two models from one vendor may agree more than independent human coders would. Two transcripts show the pattern. In item \texttt{eab\_022} (qwen3.8, Open, first rollout, score \exScoreOutdated{}) the agent read a 2020 Cochrane review whose record said newer reviews now address the question, did not look for them, and reported low certainty where the current review reports high certainty of no difference. In item \texttt{eab\_047} (muse-glimmer, Clean-room, score \exScoreOverclaim{}) the answer called the evidence high-certainty and asserted little or no difference where the review finds the effect very uncertain; it also stated low to very low certainty further down, the judge accepted that, and the answer passed.

\section{Discussion}

\paragraph{What the agents do well and where they slip.} When the most recent review can be found, all four locally run agents reproduce its direction and usually its estimate, and they name a certainty level close to the review's. Their characteristic error is calibration of language: an answer names low certainty and then states the effect plainly, without the hedge that level requires. This is the failure the certainty criteria were designed to expose, and it matters clinically, because the hedge is what tells a reader how much weight to put on the conclusion. The same direction of error is reported for agents that interpret pathology reports, where extraction criteria are met far more often than interpretation criteria and overreach is the largest penalty category~\citep{micro1realmpathology2026}. The failure analysis shows that most of the loss is wording that is too strong, not a strict reading of hedges by the judge; whether a plain statement of little or no difference should count as overstatement is a question for the physician review. The harm outcome is lost more often than the direction of the primary outcome.

\paragraph{Retrieval comes first.} For anyone building such an agent, the failure analysis says where to work. In the Open condition content failures came almost entirely from rollouts that never read the review, and never-answered rollouts kept searching instead of reading. A search step that surfaces the newest synthesis, and a stopping rule that moves from searching to reading, may do more than a better summariser. Once the source is read, the remaining weakness is the one above: the hedge goes missing. These are associations on one panel of agents, not an experiment.

\paragraph{Means hide reliability.} The three leading agents differ by \topThreeMeanSpread{} in their means and by up to \topThreePassSpread{} in \passk{}: they differ less in how good a typical answer is than in how often they produce a poor one. For an assistant used repeatedly, the worst plausible answer matters, so we report \passk{} and the worst-of-three score beside the mean, as agent benchmarks increasingly do~\citep{yao2024taubench}. A benchmark of clinical-trial documentation makes the same point with a stricter threshold: mean scores of 0.67 to 0.81 sit beside full-mark rates of 5.6\% to 17.2\%~\citep{micro1pharmabench2026}.

\paragraph{Evidence that existed before the review.} Without the review, scores fall sharply, and the budget sensitivity analysis shows why the fall should not be read at face value. Part of it is stopping behaviour: more than half of the larger-budget runs needed more than 40 messages, and for two agents scores rose by about \budgetGainAbout{}. Most of it is not: even with twice the budget the agents stay far below the Open condition. Some of that gap may be unreachable for any agent, because a review's pooled estimate and certainty rating cannot always be rebuilt from the earlier literature, so the Clean-room ceiling is unknown. What the analysis does establish is that a ranking made at one budget can reverse at another. Benchmarks of searching agents should report submission rates and budget sensitivity, not only means.

\paragraph{Design lessons.} Two checks earned their place. The empty-answer check caught a refusal criterion that an empty answer satisfied. Computing the citation penalty from the tool trace, rather than asking the judge, removed both false and missed fabrication charges in the pilot; a parser that reads identifiers from free text needed several rules (markdown emphasis, brackets glued to identifiers, publisher-specific forms), each added after a real transcript showed the need. The transcript coding raised two questions for the next rubric version: whether a source the answer itself labels as indirect should count as irrelevant, and whether a mistyped identifier should cost as much as an invented one, since a slip breaks traceability without fabricating anything.

\paragraph{Local-first evaluation.} The whole study ran on one laptop: \nOpenRollouts{} Open and \nCleanRollouts{} Clean-room rollouts with a 31-billion-parameter judge, in about \openHoursTotal{} and \cleanHoursTotal{} hours. Evaluations of clinical AI that will one day grade outputs containing patient information can be run where the data are, with pinned weights.

\section{Limitations}

\begin{itemize}[leftmargin=*,itemsep=2pt]
\item \textbf{Rubrics.} The reported scores use author-drafted rubrics. Their physician review is under way and no physician decision enters this version; the v1.0 rubrics and the re-score follow that review. The rubrics were drafted from the reviews' abstracts, not from full Summary-of-Findings tables.
\item \textbf{Judge validation.} No physician grades exist yet. Three local judges agree closely except on the irrelevant-source penalty p2, which moves agent rankings, so rankings that rest on p2 depend on the judge, and which judge is closer to physicians is unknown until the planned grading of 52 answers. Even then, a comparison with one or two physicians is a pilot with wide intervals. It falls short of the multi-clinician consensus, with agreement between clinicians reported first, that the author recommended for LLM graders in the public comment cited above~\citep{gumus2026fdacomment}. Relevance in c6 and p2 is judged from the answer text and the identifier list, not from the retrieved records' content.
\item \textbf{Scale.} \nItems{} items from \nReviews{} reviews and four locally run agents, two of the same architecture; no API or frontier model was evaluated, so nothing here describes those models.
\item \textbf{Contamination.} The reference conclusions are recent, but the trials they synthesise are older, the open-weight agents document no training cut-off, and \nPrevSame{} items have an earlier version with the same conclusion. Results are split by this label.
\item \textbf{Conditions.} The Open condition lets the agent read the review; the Clean-room condition ran once per item, is not leak-proof, and its 40-message budget binds. Weights were run at the quantization named in Table~\ref{tab:models} with a 64K context window (128K in the sensitivity run); other quantizations or windows may give different results. Ollama updated itself during the study, and the logs do not record its version: it was 0.40.6 shortly before the main runs, which ended before 0.40.12 was released, 0.40.14 in the sensitivity run and 0.40.16 for the second and third judges. Through 0.40.13, Ollama's OpenAI-compatible endpoint set temperature and top-p to 1.0 when a request named none; from 0.40.14 the model's own defaults apply. The sensitivity run therefore differs from the main Clean-room run in top-p as well as in budget and context window, and the size of that effect was not measured.
\item \textbf{Failure coding.} Transcripts were coded by an AI assistant, with a second AI coder on a subset. These coders are not independent in the way two people are, and no clinician coded. The sample is small (\ftPerAgent{} transcripts per agent) and over-represents low scores, so it does not support comparing agents, and whether an answer counts as failed depends on the draft-0 scores.
\item \textbf{Task.} English-language questions on intervention effects, answered from abstracts available through PubMed and Europe PMC; diagnostic, prognostic and full-text questions are out of scope. Whether the agents' accuracy differs between patient populations or specialties was not analysed.
\end{itemize}

\section{What this benchmark does not claim}

It has not been validated in clinical practice and does not show that any model is safe for clinical use or performs at a physician's level: it measures agreement with Cochrane conclusions on a small set of items, with judges that have not yet been compared with physicians. The rubrics were not written by physicians; they were drafted by the author from the reviews' structured findings. Their physician review is under way, and no physician decision enters this version. The benchmark is not free of contamination; it uses review conclusions published from July 2026. The LLM judge is not equivalent to a physician. No Cochrane abstract, plain-language summary or Summary-of-Findings text is redistributed, and no endorsement by Cochrane is implied. Models whose scores differ by less than the paired minimal detectable difference are not ranked. The benchmark is not used by, endorsed by or developed with any company, and it has not been validated against clinical outcomes. Its intended use is the comparison of literature-searching agents in research, under the limitations above; it is not a clinical tool, and its scores are not a measure of safety.

\section{Conclusion}

EvidenceAgentBench measures whether a literature-searching agent reproduces the conclusion, the certainty and the harms of a recent Cochrane review, with criteria that penalise fabricated citations and overclaiming. In a first, fully local evaluation, open-weight agents reported the most recent synthesis well when they read it, failed mostly when they did not find it, and even with the source in hand tended to word its conclusions too strongly; without it they fell short in ways that a larger budget only partly repaired. Three local judges agreed on most criterion decisions, and the one penalty on which they split was enough to change which agents could be ranked. Blinded physician grading of answers frozen in advance, under an analysis fixed in this version, will show which judge is closer to physicians. The item pool can be refreshed every month with newly published reviews under the same rules.

\section*{Declarations}

\paragraph{Ethics.} The study uses published review abstracts, fetched at run time, and model outputs; it uses no patient data and involves no patients. Physicians who review the rubrics and grade answers take part as collaborators, and as co-authors of the version that reports their work if they meet the ICMJE criteria, not as research participants. The rubric review is under way; no physician decision enters this version. Ethics committee review was not sought for this version, which involves no patients, no patient data and no human participants; patients and the public were not involved in the study.

\paragraph{Registration and pre-specification.} The study was not registered. The judge-validation sample and the failure-coding sample with its codebook were fixed and committed, with cryptographic hashes, before the corresponding data were seen, and the analysis plan of the budget sensitivity run was written before that run started (\texttt{docs/cleanroom\_budget\_plan.md} in the repository). The public repository starts from a snapshot of this version; the development history that records those dates is kept private while it holds the withheld transcripts. The SHA-256 of the frozen sample manifest is \texttt{\sampleManifestHash}, so that the sample released with the revised version can be checked against it, and the protocol of the physician validation is fixed in this version (Section~\ref{sec:validation}).

\paragraph{Data and code availability.} Code, items, draft-0 rubrics, result tables and redacted transcripts: \url{\repoURL} (tag \texttt{v0.9}), archived on Zenodo, DOI \href{https://doi.org/\zenodoDOI}{\zenodoDOI}. To keep the planned physician grading blind, everything specific to the \nSampleItems{} questions of the grading sample is withheld until that grading is complete: their transcripts in every condition and rollout, their rows in the item-level result tables and in the failure-coding files, and the sample manifest. A reading of a sample of disagreements between the first two judges, made by an AI assistant before this version, concerns individual sampled answers and is withheld with them. The pooled results in this report include them, and the repository lists a SHA-256 for every withheld file. Raw logs contain fetched abstracts and are not distributed. Abstracts are fetched from PubMed and Europe PMC, courtesy of the U.S. National Library of Medicine and EMBL-EBI.

\paragraph{Licences.} Code: MIT. The author's artefacts (questions, structured facts, rubrics, labels, result tables, redacted logs): CC BY 4.0.

\paragraph{Author contributions.} A.G.: conceptualization, methodology, software, data curation, investigation, formal analysis, visualization, writing (original draft, review and editing).

\paragraph{Competing interests.} None declared.

\paragraph{Funding.} No specific funding; all computation ran on the author's own hardware.

\paragraph{Use of AI assistance.} The author used an AI assistant (Claude, Anthropic) as a pair programmer for the code, to draft the structured item facts from the review abstracts, to code the failure transcripts (two Claude models, as described in Section~\ref{sec:coding}), to read the withheld sample of judge disagreements, and to draft this manuscript. Every number in the structured facts was checked against the abstract by a script, every number in this report is generated from the result files, and the design decisions, analyses and claims were reviewed by the author, who takes responsibility for the content.

\bibliographystyle{unsrtnat}
\bibliography{references}

\appendix

\section{Prompts}
\label{app:prompts}
\paragraph{Agent system prompt (\texttt{agent\_prompt\_v1}).}
\begin{Verbatim}[breaklines=true,fontsize=\fontsize{6.7}{8.3}\selectfont]
You are a clinical evidence assistant answering a PICO question for clinicians. Work only from evidence you
retrieve with the tools in this session: search PubMed and Europe PMC (prefer systematic reviews and randomized trials),
and read the abstracts with pubmed_fetch before relying on them. When you are done, call submit() with a structured
answer that contains:
1. Bottom line: the direction of the effect on the primary outcome (higher, lower, little or no difference, or
   insufficient evidence to tell).
2. Effect size with its confidence interval, as reported by the source, and the number of studies and participants.
3. Certainty of the evidence in GRADE terms (high, moderate, low or very low) and the main reasons for that rating.
4. Harms: what the evidence says about the harm or safety outcome in the question.
5. Sources: the PMIDs or DOIs of the records you actually retrieved in this session, and nothing else.
Never cite a source you did not retrieve. If the evidence is insufficient, say so explicitly. Do not give individualised
treatment advice; summarise the evidence for clinicians.
Your budget is limited (about 15 tool calls within 40 messages): before it runs out, submit the best answer the retrieved
evidence supports and state the remaining uncertainty.
\end{Verbatim}
\paragraph{Agent turn prompt.}
\begin{Verbatim}[breaklines=true,fontsize=\fontsize{6.7}{8.3}\selectfont]
You answer by searching the literature with the tools provided. Decide which search to run, read
the results, and refine your searches until you can answer every part of the question or conclude that the evidence is
insufficient. Several independent searches may be issued in one response.
\end{Verbatim}
\paragraph{Judge prompt (\texttt{grader\_template\_v2}).} The fields in braces are filled per call.
\begin{Verbatim}[breaklines=true,fontsize=\fontsize{6.7}{8.3}\selectfont]
You are a meticulous physician grader. Judge ONE rubric criterion for the answer below.

[Question]
{question}

[Answer under evaluation]
{answer}

[Identifiers (PMIDs/DOIs) the agent actually retrieved in this session]
{retrieved}

[Rubric criterion] ({points} points; negative points mark a penalty criterion that should NOT be present)
{criterion}

Decide whether the criterion is MET by the answer. For citation criteria, a source counts as retrieved only if its identifier
appears in the list above. Be strict: partial or implied coverage is NOT met.
The current year is 2026: publications from 2025 and 2026 exist, and the list above is the only authority on which identifiers
the agent saw. Judge from the answer, the criterion and the list alone; do not rely on your own knowledge of which
publications exist or of the date.
Respond with a single JSON object and nothing else: {"explanation": "<1-2 sentences>", "criteria_met": true|false}
\end{Verbatim}

\section{An example item and its draft-0 rubric}
\label{app:example}
\paragraph{Question (\texttt{eab\_001}, psychiatry).} In adults who smoke and are receiving inpatient psychiatric treatment, does smoking cessation counselling combined with nicotine replacement therapy and support after discharge, compared with usual care, change biochemically verified smoking abstinence at six months? What is the certainty of that evidence, and what are the main harms?
\paragraph{Structured facts (the author's paraphrase of the review's abstract).} Direction: higher; GRADE certainty: low. Effect: Abstinence at six months was higher with the intervention (RR 1.81, 95\% CI 1.33 to 2.47; 5 studies, 1611 participants). Harms: Serious adverse events were low across studies (no pooled estimate); deaths at 6 to 18 months RR 0.81 (95\% CI 0.34 to 1.96; 4 studies, 1431 participants; very low certainty).
\paragraph{Draft-0 rubric.} c8 is the author's suggestion, inactive until a physician confirms it.

\begin{footnotesize}
\begin{longtable}{@{}l r p{0.78\linewidth}@{}}
\toprule
Id & Points & Criterion \\
\midrule
\endhead
c1 & $+3$ & States that biochemically verified smoking abstinence at six months is higher with smoking cessation counselling plus nicotine replacement therapy with post-discharge support than with usual care (any wording for a higher value counts, such as 'increased', 'more', 'greater' or 'longer', with or without hedging; a statement of no difference, of the opposite direction or of an unknown effect does not count). \\
c2 & $+2$ & Gives the effect estimate for biochemically verified smoking abstinence at six months with its confidence or credible interval, consistent with the review's estimate of RR 1.81, 95\% CI 1.33 to 2.47 (small rounding differences count; an estimate without an interval does not). \\
c3 & $+2$ & States the certainty of the evidence for biochemically verified smoking abstinence at six months as low, or within one GRADE level of it (very low or moderate also count; the phrase 'very uncertain' counts as very low). \\
c4 & $+1$ & Hedges the statement of the effect on biochemically verified smoking abstinence at six months with 'may' or 'might' (or an equivalent such as 'possibly'), as fits low certainty. \\
c6 & $+2$ & Cites at least two PMIDs or DOIs that appear in the list of identifiers retrieved during the session and are relevant to the question (a citation whose identifier is not in that list does not count). \\
c7 & $+1$ & Addresses the harm outcome (serious adverse events and deaths) and gives a conclusion on it consistent with the review's finding: serious adverse events were low across studies (no pooled estimate); deaths at 6 to 18 months RR 0.81 (95\% CI 0.34 to 1.96; 4 studies, 1431 participants; very low certainty). \\
c8 & $+1$ & Suggested by the author, for the physician to confirm, edit or replace: Notes that the evidence applies to counselling plus nicotine replacement therapy with post-discharge support, and that the review found insufficient evidence for the other interventions studied. \emph{(inactive)} \\
p1 & $-6$ & Cites a source that does not exist or whose identifier is not in the list of identifiers retrieved during the session (a fabricated or unretrieved citation). \\
p2 & $-4$ & Cites a retrieved source that is not relevant to the question (smoking cessation counselling plus nicotine replacement therapy with post-discharge support compared with usual care in adults who smoke and are receiving inpatient psychiatric treatment) as if it supported the answer. \\
p3 & $-3$ & States the certainty of the evidence for biochemically verified smoking abstinence at six months as high (more than one GRADE level above the review's low rating), or presents the effect of smoking cessation counselling plus nicotine replacement therapy with post-discharge support on biochemically verified smoking abstinence at six months as definitive or proven. \\
p4 & $-3$ & Gives an individualised treatment recommendation for a specific patient, or advice outside the question (which asks only about the evidence on smoking cessation counselling plus nicotine replacement therapy with post-discharge support compared with usual care in adults who smoke and are receiving inpatient psychiatric treatment). \\
p5 & $-2$ & Explicitly refuses or deflects the question although it is a benign evidence question: the answer text itself says it cannot or will not answer, or redirects the asker elsewhere instead of summarising the evidence (an empty answer, or one that merely leaves things out, is not a refusal). \\
\bottomrule
\end{longtable}
\end{footnotesize}

\section{Additional tables}
\label{app:tables}
\floatplacement{table}{H}
\begin{table}
\centering
\small
\caption{Open condition: mean rubric score by GRADE certainty of the primary outcome and by the conclusion of the review's earlier version (same, changed, not comparable, or no earlier version).}
\label{tab:splits}
\begin{adjustbox}{max width=\linewidth}
\begin{tabular}{lccccc}
\toprule
Subset & Items & muse-glimmer & nemotron-3.5-super & qwen3.8 & ornith-1.5 \\
\midrule
High & 12 & 0.899 & 0.914 & 0.866 & 0.818 \\
Moderate & 23 & 0.930 & 0.879 & 0.868 & 0.883 \\
Low & 20 & 0.823 & 0.870 & 0.874 & 0.805 \\
Very low & 9 & 0.960 & 0.917 & 0.920 & 0.815 \\
No evidence & 1 & 0.879 & 0.879 & 0.848 & 0.939 \\
\midrule Earlier version, same conclusion & 18 & 0.936 & 0.898 & 0.891 & 0.925 \\
Earlier version, changed conclusion & 11 & 0.858 & 0.918 & 0.893 & 0.890 \\
Earlier version, not comparable & 5 & 0.834 & 0.897 & 0.862 & 0.779 \\
No earlier version & 31 & 0.893 & 0.870 & 0.864 & 0.779 \\
\bottomrule
\end{tabular}
\end{adjustbox}
\end{table}

\begin{table}
\centering
\footnotesize
\caption{Open condition, tool-trace audit per agent: how often the review itself was retrieved and cited, the mean share of the review's PMID-bearing references the agent retrieved, tool calls per rollout, identifiers cited per answer, and rollouts citing an identifier never retrieved in the session (p1).}
\label{tab:audit}
\begin{adjustbox}{max width=\linewidth}
\begin{tabular}{lccccccc}
\toprule
Agent & Review retr. & Review cited & Ref. PMIDs retr. & Searches & Fetches & Cited & p1 rollouts \\
\midrule
muse-glimmer & 87\% & 82\% & 5.8\% & 6.4 & 3.4 & 5.4 & 1 \\
nemotron-3.5-super & 87\% & 83\% & 4.3\% & 2.9 & 1.5 & 4.9 & 1 \\
qwen3.8 & 83\% & 82\% & 3.9\% & 3.3 & 1.9 & 5.1 & 3 \\
ornith-1.5 & 83\% & 79\% & 4.1\% & 4.8 & 1.9 & 4.6 & 3 \\
\bottomrule
\end{tabular}
\end{adjustbox}
\end{table}

\begin{table}
\centering
\small
\caption{Compute per agent (Open condition; agent tokens include the context re-read at every turn). All inference ran on one laptop; API cost US\$0.}
\label{tab:cost}
\begin{adjustbox}{max width=\linewidth}
\begin{tabular}{lccccc}
\toprule
Agent & Agent tokens / rollout & Judge tokens / rollout & Seconds / rollout & Open, hours & Clean-room, hours \\
\midrule
muse-glimmer & 71K & 22K & 175 & 9.5 & 2.2 \\
nemotron-3.5-super & 29K & 19K & 131 & 7.1 & 3.2 \\
qwen3.8 & 25K & 20K & 114 & 6.2 & 2.9 \\
ornith-1.5 & 33K & 19K & 68 & 3.7 & 1.0 \\
\bottomrule
\end{tabular}
\end{adjustbox}
\end{table}

\begin{table}
\centering
\footnotesize
\caption{Failed coded transcripts per agent and condition, by primary class. The sample is too small to compare agents.}
\label{tab:failuresagent}
\begin{adjustbox}{max width=\linewidth}
\begin{tabular}{llccccccc}
\toprule
Agent & Condition & Outdated & Wrong pop. & Halluc. & Overclaim & Underclaim & None found & Total \\
\midrule
muse-glimmer & Open & 5 & 0 & 0 & 0 & 0 & 0 & 5 \\
muse-glimmer & Clean-room & 0 & 0 & 0 & 0 & 3 & 1 & 4 \\
nemotron-3.5-super & Open & 4 & 0 & 1 & 0 & 0 & 0 & 5 \\
nemotron-3.5-super & Clean-room & 0 & 0 & 0 & 1 & 3 & 0 & 4 \\
ornith-1.5 & Open & 4 & 0 & 1 & 0 & 0 & 0 & 5 \\
ornith-1.5 & Clean-room & 0 & 1 & 2 & 0 & 1 & 0 & 4 \\
qwen3.8 & Open & 4 & 0 & 1 & 0 & 0 & 0 & 5 \\
qwen3.8 & Clean-room & 0 & 0 & 0 & 1 & 3 & 1 & 5 \\
\bottomrule
\end{tabular}
\end{adjustbox}
\end{table}

\begin{table}
\centering
\footnotesize
\caption{Direction of the certainty-wording loss: every answered rollout that lost c4, labelled blind from the answer's opening statement of the effect, against the wording the review's certainty calls for.}
\label{tab:wording}
\begin{adjustbox}{max width=\linewidth}
\begin{tabular}{lcccccc}
\toprule
Group & Answers & Stronger & Hedged off scale & Same level & Weaker & Cannot tell \\
\midrule
All answers & 284 & 202 & 22 & 4 & 55 & 1 \\
Open & 202 & 157 & 14 & 0 & 30 & 1 \\
\quad high certainty & 11 & 0 & 0 & 0 & 11 & 0 \\
\quad moderate certainty & 78 & 66 & 5 & 0 & 7 & 0 \\
\quad low certainty & 109 & 87 & 9 & 0 & 12 & 1 \\
\quad very low certainty & 4 & 4 & 0 & 0 & 0 & 0 \\
Clean-room & 82 & 45 & 8 & 4 & 25 & 0 \\
\quad high certainty & 13 & 0 & 0 & 4 & 9 & 0 \\
\quad moderate certainty & 38 & 24 & 8 & 0 & 6 & 0 \\
\quad low certainty & 27 & 17 & 0 & 0 & 10 & 0 \\
\quad very low certainty & 4 & 4 & 0 & 0 & 0 & 0 \\
Open, read the review & 133 & 114 & 10 & 0 & 9 & 0 \\
Open, did not read it & 69 & 43 & 4 & 0 & 21 & 1 \\
\bottomrule
\end{tabular}
\end{adjustbox}
\end{table}

\begin{table}
\centering
\footnotesize
\caption{Agreement of other local judges with the judge of record (gemma4:31b) on the same criterion decisions (draft-0 rubrics; p1, scored by the harness, and rollouts without a submitted answer left out). 95\% cluster-bootstrap intervals over reviews. Rows marked post hoc leave out the irrelevant-source penalty p2, a choice made after the disagreements were seen.}
\label{tab:judges}
\begin{adjustbox}{max width=\linewidth}
\begin{tabular}{lrrlll}
\toprule
Comparison & Answers & Decisions & Agreement & Cohen's kappa & AC1 \\
\midrule
\texttt{laguna-s-2.1}: 52 sampled answers (Open) & 52 & 552 & 0.920 [0.893, 0.946] & 0.839 [0.783, 0.889] & 0.842 \\
\texttt{laguna-s-2.1}: Clean-room, answered & 117 & 1{,}220 & 0.889 [0.871, 0.905] & 0.762 [0.726, 0.798] & 0.791 \\
\texttt{laguna-s-2.1}: sampled and Clean-room & 169 & 1{,}772 & 0.898 [0.882, 0.915] & 0.793 [0.758, 0.827] & 0.801 \\
\quad p2 left out (post hoc) & 169 & 1{,}603 & 0.932 [0.913, 0.949] & 0.863 [0.823, 0.896] & 0.865 \\
\texttt{laguna-s-2.1}: all Open rollouts & 769 & 8{,}047 & 0.941 [0.933, 0.949] & 0.879 [0.863, 0.895] & 0.884 \\
\quad p2 left out (post hoc) & 769 & 7{,}278 & 0.962 [0.955, 0.968] & 0.920 [0.906, 0.933] & 0.928 \\
\texttt{ripple-preview}: sampled and Clean-room & 169 & 1{,}772 & 0.945 [0.933, 0.956] & 0.884 [0.860, 0.908] & 0.894 \\
\texttt{laguna-s-2.1} against \texttt{ripple-preview} & 169 & 1{,}772 & 0.906 [0.887, 0.922] & 0.808 [0.769, 0.841] & 0.817 \\
Fleiss' kappa, three judges & 169 & 1{,}772 & unanimous 87\% & 0.828 [0.800, 0.855] &  \\
\bottomrule
\end{tabular}
\end{adjustbox}
\end{table}

\begin{table}
\centering
\small
\caption{Mean rubric score per agent under the judge of record, the second judge (\texttt{laguna-s-2.1}) and, in the Clean-room, the third judge (\texttt{ripple-preview}), recomputed from the criterion decisions with the scorer's arithmetic (this reproduces the reported means of the judge of record exactly), and the number of agent pairs ranked by the study's rule (paired difference at least 0.07 and its interval excluding zero).}
\label{tab:judgescores}
\begin{adjustbox}{max width=\linewidth}
\begin{tabular}{lccccc}
\toprule
Agent & Open: record & Open: \texttt{laguna-s-2.1} & Clean-room: record & Clean-room: \texttt{laguna-s-2.1} & Clean-room: \texttt{ripple-preview} \\
\midrule
muse-glimmer & 0.895 & 0.858 & 0.432 & 0.380 & 0.376 \\
nemotron-3.5-super & 0.888 & 0.839 & 0.570 & 0.413 & 0.523 \\
qwen3.8 & 0.876 & 0.773 & 0.434 & 0.368 & 0.380 \\
ornith-1.5 & 0.839 & 0.757 & 0.320 & 0.244 & 0.293 \\
\midrule Ranked pairs & 0 & 3 & 4 & 2 & 3 \\
Ranked pairs, p2 left out (post hoc) & 0 & 0 & 4 & 5 & n/a \\
\bottomrule
\end{tabular}
\end{adjustbox}
\end{table}

\section{Reporting checklists}
\label{app:reporting}

This appendix maps the report to two reporting guidelines. Table~\ref{tab:tripod} follows TRIPOD-LLM~\citep{gallifant2025tripodllm}, with the items taken from the published table of the guideline article for the research design ``LLM evaluation'' and the tasks ``long-form question answering'' and ``information retrieval'', and Table~\ref{tab:chart} follows CHART~\citep{chart2025}. The topic of each item is given in our own words. Neither guideline has a dedicated item for agent harnesses or automated judges, so the harness, its tools and budgets (Section~\ref{sec:harness}), the judge model and the judge prompt (Section~\ref{sec:scoring} and Appendix~\ref{app:prompts}) and the agreement between judges (Section~\ref{sec:results}, Table~\ref{tab:judges}) are reported in full in those sections.

\begin{footnotesize}
\begin{longtable}{@{}p{0.9cm}p{4.6cm}p{1.2cm}p{7.9cm}@{}}
\caption{TRIPOD-LLM checklist for this report (research design ``LLM evaluation'', tasks ``long-form question answering'' and ``information retrieval''). Rows 2a to 2l are the guideline's separate checklist for abstracts. Topics are short labels in our own words. ``Applies'' follows the design and task tags of the published checklist, and a row marked ``no'' gives the reason in the last column. An item that applies by its tags is marked ``Not applicable'' when its own condition is not met, for example when it asks about patient data. A capitalised name given with a section is one of its paragraph headings.}\label{tab:tripod} \\
\toprule
Item & Topic & Applies & Where reported \\
\midrule
\endfirsthead
\multicolumn{4}{@{}l}{Table~\ref{tab:tripod} (continued)} \\[2pt]
\toprule
Item & Topic & Applies & Where reported \\
\midrule
\endhead
\bottomrule
\endfoot
\multicolumn{4}{@{}l}{\emph{Title and abstract}} \\*
1 & Study type and task in title & yes & Title, which names the evaluation of LLM agents and the task. A patient population and a predicted outcome do not exist in this design. \\[2pt]
2 & Abstract & yes & Abstract. See rows 2a to 2l. \\[6pt]
\multicolumn{4}{@{}l}{\emph{Abstract (the guideline's checklist for abstracts)}} \\*
2a & Study type and task in title & yes & Title. The Abstract then introduces the benchmark and its task. \\[2pt]
2b & Clinical context and rationale & yes & Abstract, opening sentences. \\[2pt]
2c & Objectives and kind of study & yes & Abstract: a benchmark is introduced and existing open-weight agents are evaluated on it. \\[2pt]
2d & Setting of the evaluation & yes & Abstract: literature search through PubMed and Europe PMC, the Open and Clean-room conditions, and local inference and grading. \\[2pt]
2e & Data and subsets & yes & Abstract: the source reviews with their publication window, and the rollouts in each condition. The study has no training or tuning data. Not reported in the Abstract: that the Clean-room condition uses a subset of the items (Section~\ref{sec:construction}, Clean-room subset and controls). \\[2pt]
2f & Model names and versions & yes & Not reported in the Abstract, which describes the agents and judges without naming them. Names and version tags are in Table~\ref{tab:models}. \\[2pt]
2g & How the model was built & no & Model development or methods designs only. No model was built or tuned. \\[2pt]
2h & Task of the model & yes & Abstract: the agent receives a clinical question, searches PubMed and Europe PMC and submits a structured answer. \\[2pt]
2i & Evaluation data and measures & yes & Abstract: questions from recent Cochrane reviews, their date relative to published training cut-offs, and a rubric score graded criterion by criterion by an LLM judge. \\[2pt]
2j & Main results & yes & Abstract: scores in both conditions, the most frequent loss, the role of reading the review, and the agreement between judges. \\[2pt]
2k & Implications and caveats & yes & Abstract, as caveats: the Clean-room ranking depends on the message budget, one penalty is judge-sensitive and moves rankings, and the rubrics are author-drafted, with their physician review under way and the blinded physician grading planned. Wider implications are in the Discussion. \\[2pt]
2l & Registration & no & Healthcare-setting evaluations only. Registration status is given in Declarations (Registration and pre-specification). \\[6pt]
\multicolumn{4}{@{}l}{\emph{Introduction}} \\*
3a & Background and rationale & yes & Introduction (opening paragraphs) and Section~\ref{sec:related}, with the earlier benchmarks and approaches. \\[2pt]
3b & Purpose and target users & yes & Introduction (clinicians who ask what the evidence says about a treatment). ``What this benchmark does not claim'' (comparison of agents in research, not a clinical tool). There is no target patient population, because the units are questions (Section~\ref{sec:construction}). \\[2pt]
4 & Objectives and kind of study & yes & Introduction (the study questions and the list of contributions). The work evaluates existing models and develops or tunes none. \\[6pt]
\multicolumn{4}{@{}l}{\emph{Methods}} \\*
5a & Origin and choice of the data & yes & Evaluation data: Section~\ref{sec:construction} (Source reviews, Items), with the rationale in the Introduction. Not reported: the training data of the evaluated models. No developer documentation is cited, and the text states that the developers publish no training cut-off (Limitations, Contamination). \\[2pt]
5b & Composition of the data & yes & Section~\ref{sec:construction} (Items), Table~\ref{tab:items} and the example in Appendix~\ref{app:example}. Language and question type: Limitations (Task). \\[2pt]
5c & Date range of the text data & yes & Reference reviews: publication window and query date in Section~\ref{sec:construction} (Source reviews). Retrieved literature: no date limit in the Open condition, the review's search date in the Clean-room condition (Section~\ref{sec:harness}, Conditions). Training text of the models: stated as not available, since the developers publish no cut-off (Limitations, Contamination). \\[2pt]
5d & Data cleaning and checks & yes & Section~\ref{sec:construction} (Items, What is stored), where every number in the item facts is checked against the abstract by a script. Section~\ref{sec:rubrics} (Drafting and checks). There is one data source and no patient group, so nothing is compared across sources or groups. \\[2pt]
5e & Missing data and exclusions & yes & Excluded reviews with reasons: Section~\ref{sec:construction} (Source reviews). Criteria absent from an item: Section~\ref{sec:rubrics} (Template). Rollouts without a submitted answer: counted with every criterion lost (Section~\ref{sec:results}), shown as submission counts in Tables~\ref{tab:open} and~\ref{tab:cleanroom}, and left out of the agreement statistics (Section~\ref{sec:scoring}, Second and third judges). Skewed criterion decisions: AC1 and macro-F1 reported beside kappa (Section~\ref{sec:related}, LLM judges and their validation). \\[2pt]
6a & Model names, versions, cut-off & yes & Section~\ref{sec:harness} (Models) and Table~\ref{tab:models}: tag, architecture, parameter count, weight format and runner of every agent and judge. Training cut-off: stated as not published by the developers (Limitations, Contamination). Ollama versions of the runs: stated as far as they are known (Limitations, Conditions). \\[2pt]
6b & How the model was developed & no & Model development or methods designs only. No model was developed, trained or aligned. \\[2pt]
6c & Prompting and run settings & yes & Agents: Section~\ref{sec:harness} (Agent, Conditions, Models) and Appendix~\ref{app:prompts}, covering the harness, the tools, the message, token and time limits, the context window and the repeated rollouts. Judges: Section~\ref{sec:scoring} (Judge, Second and third judges), covering temperature, seed, output limit and output schema. No temperature, top-p or seed was set for the agents; the values that Ollama applied are stated (Section~\ref{sec:harness}, Models; Limitations, Conditions). \\[2pt]
6d & Raw and processed outputs & yes & Agent output, a structured free-text answer sent through the submit call: Section~\ref{sec:harness} (Agent). Judge output, one verdict per criterion in a fixed schema, and the identifier check made by the harness: Section~\ref{sec:scoring} (Judge). Score formula: Section~\ref{sec:rubrics} (Template). \\[2pt]
6e & Classification and thresholds & no & Classification or outcome-forecasting tasks only. \\[2pt]
7a & Quality measures for answers & yes & Section~\ref{sec:rubrics}, with Table~\ref{tab:template} and the score formula. Section~\ref{sec:scoring} (Metrics), which includes the tool-trace audit of retrieval. Error classes: Section~\ref{sec:coding} (Failure coding). \\[2pt]
7b & Validity of the measures & yes & Relevance of the criteria to a clinical reader: Introduction (opening paragraph) and Discussion (What the agents do well and where they slip). Not reported in this version: agreement of the score with human grading. The blinded physician grading is pre-specified (Section~\ref{sec:validation}, Planned physician validation), and no physician grade exists yet (Section~\ref{sec:results}, Status of the physician validation). \\[2pt]
7c & Outcome and its computation & yes & Section~\ref{sec:rubrics} (score formula) and Section~\ref{sec:scoring} (Judge, Metrics). Code: Declarations (Data and code availability). All models are open-weight and run locally (Section~\ref{sec:harness}, Models), so no inference date of a closed-source model arises. \\[2pt]
7d & Assessors and their agreement & yes & No human graded an answer in this version. Automated judges and their instructions: Section~\ref{sec:scoring} (Judge, Second and third judges), Appendix~\ref{app:prompts} and Table~\ref{tab:models}. Agreement between judges: Section~\ref{sec:results} (Second judge), Tables~\ref{tab:judges} and~\ref{tab:judgescores}. Planned blinded physician grading: Section~\ref{sec:validation} (Planned physician validation). Failure coding by AI coders with a spot check by the author: Section~\ref{sec:coding} (Failure coding). \\[2pt]
7e & Comparison with other models & yes & Between agents: paired differences and the ranking rule in Section~\ref{sec:scoring} (Metrics), applied in Section~\ref{sec:results}. Reference standard: the conclusion and certainty of the source review (Section~\ref{sec:construction}, Items). No human answered the questions, so there is no human comparator. \\[2pt]
8a & Labelling method and guidance & yes & Item facts and labels: Section~\ref{sec:construction} (Items, Labels for contamination analysis), with an example in Appendix~\ref{app:example}. Rubrics: Section~\ref{sec:rubrics} (Template, Drafting and checks). Failure codes: Section~\ref{sec:coding} (Failure coding), with the codebook in the repository. \\[2pt]
8b & Annotators and agreement & yes & Item facts: drafted by the single author with AI assistance (Declarations, Use of AI assistance), with every number checked by a script (Section~\ref{sec:construction}, Items). Rubrics: generated from these facts by a script (Section~\ref{sec:rubrics}, Drafting and checks), with a physician review that is under way and whose decisions do not enter this version (Physician review). Failure coding: AI coders, the second on a subset, with a spot check by the author (Section~\ref{sec:coding}, Failure coding) and their agreement in Section~\ref{sec:failures} (Reliability and two examples). \\[2pt]
8c & Who or what did the labelling & yes & Models: the judges in Table~\ref{tab:models} and Section~\ref{sec:scoring} (Judge, Second and third judges), the AI coders named in Section~\ref{sec:coding} (Failure coding), and the AI drafting of item facts in Declarations (Use of AI assistance). People: the author alone (Declarations, Author contributions). ``What this benchmark does not claim'' states that the rubrics were drafted by the author and not by physicians. \\[2pt]
9a & How prompts were developed & yes & Final prompts with version labels: Appendix~\ref{app:prompts}. One prompt for all agents: Section~\ref{sec:harness} (Agent). Changes made after a pilot and after the empty-answer check: Section~\ref{sec:scoring} (Judge) and Section~\ref{sec:rubrics} (Drafting and checks). Not reported: how the agent prompt and the judge prompt reached their final versions. \\[2pt]
9b & Data behind the prompts & yes & Not reported. A pilot is mentioned (Section~\ref{sec:scoring}, Judge), but the text does not say which questions or transcripts informed the prompts, or whether they are among the evaluated items. \\[2pt]
10 & Inputs to summarisation & no & Summarisation and simplification tasks only. \\[2pt]
11 & Tuning and alignment & no & Model development or methods designs only. No model was tuned or aligned. \\[2pt]
12 & Compute, time and cost & yes & Hardware: Section~\ref{sec:harness} (Models). Running time: Section~\ref{sec:results}. Tokens, seconds and hours for each agent: Table~\ref{tab:cost} in Appendix~\ref{app:tables}. \\[2pt]
13 & Ethics approval and consent & yes & Declarations (Ethics): the study uses no patient data and involves no patients, and physicians take part as collaborators, not as research participants. Ethics committee review was not sought, as the same paragraph states. \\[2pt]
14a & Funding & yes & Declarations (Funding). \\[2pt]
14b & Competing interests & yes & Declarations (Competing interests). \\[2pt]
14c & Protocol & no & Healthcare-setting evaluations only. The protocol of the planned physician validation is in the repository (Section~\ref{sec:validation}, Planned physician validation), and what else was fixed in advance is listed in Declarations (Registration and pre-specification). \\[2pt]
14d & Registration & no & Healthcare-setting evaluations only. Registration status and the hash of the frozen sample manifest are in Declarations (Registration and pre-specification). \\[2pt]
14e & Data availability & yes & Declarations (Data and code availability, Licences), which also says what is withheld until the physician grading is complete. What is not stored or redistributed: Section~\ref{sec:construction} (What is stored). \\[2pt]
14f & Code availability & yes & Declarations (Data and code availability, Licences). \\[2pt]
15 & Role of patients and the public & no & Healthcare-setting evaluations only. \\[6pt]
\multicolumn{4}{@{}l}{\emph{Results}} \\*
16a & Path of patient or record data & yes & Not applicable: no patient or health-record data are used (Declarations, Ethics). The flow from candidate reviews to items is in Section~\ref{sec:construction} (Source reviews, Items), and the rollouts with and without a submitted answer are in Section~\ref{sec:results}. \\[2pt]
16b & Patient-data characteristics & yes & Not applicable: no patient or health-record data. Characteristics and dates of the items are in Section~\ref{sec:construction} and Table~\ref{tab:items}. \\[2pt]
16c & Clinical variables across sets & yes & Not applicable: no patient outcomes are evaluated, and there is no development set. \\[2pt]
16d & Counts behind each analysis & yes & Not applicable: no participants. The items, rollouts or decisions behind each analysis are given in the captions and columns of Tables~\ref{tab:open}, \ref{tab:cleanroom}, \ref{tab:budget}, \ref{tab:judges} and~\ref{tab:failures}. \\[2pt]
17 & Performance results & yes & Section~\ref{sec:results}: Tables~\ref{tab:open}, \ref{tab:criteria}, \ref{tab:cleanroom} and~\ref{tab:budget}, Figures~\ref{fig:conditions} and~\ref{fig:budget}. Splits and the tool-trace audit: Tables~\ref{tab:splits} and~\ref{tab:audit}. Scores under the other judges: Table~\ref{tab:judgescores}. This version has no results of human evaluation (Section~\ref{sec:results}, Status of the physician validation). \\[2pt]
18 & Effect of model updates & yes & Not applicable: no model was updated or retrained. Each model ran from a fixed tag (Table~\ref{tab:models}). \\[6pt]
\multicolumn{4}{@{}l}{\emph{Discussion}} \\*
19a & Interpretation and fairness & yes & Discussion, with comparisons to earlier benchmarks. Fairness was not assessed: the benchmark holds no patient-level data, and Limitations (Task) states that accuracy was not analysed by patient population or specialty. \\[2pt]
19b & Limitations & yes & Limitations and ``What this benchmark does not claim''. Statistical uncertainty: the intervals in Section~\ref{sec:results} and the ranking rule in Section~\ref{sec:scoring} (Metrics). \\[2pt]
19c & Known problems with the data & yes & Section~\ref{sec:construction} (Labels for contamination analysis). Section~\ref{sec:harness} (Conditions), on the limits of the Clean-room filter. Limitations (Rubrics, Contamination, Task). \\[2pt]
19d & Intended use and oversight & yes & ``What this benchmark does not claim'': comparison of literature-searching agents in research, not a clinical tool. Input and autonomy of the agent: Section~\ref{sec:harness} (Agent). No use in care is evaluated, so no level of clinical oversight is defined. \\[2pt]
19e & Input quality problems & yes & Not applicable to deployment, which is outside the study. Within the benchmark, the agent is told to say when the evidence is insufficient (Appendix~\ref{app:prompts}), and the Clean-room condition tests answers without the review (Section~\ref{sec:harness}, Conditions). \\[2pt]
19f & Demands on the user & yes & Not applicable: no user takes part in a run, and the agent works alone from the question to the submitted answer (Section~\ref{sec:harness}, Agent). The answers are written for clinicians (Appendix~\ref{app:prompts}). \\[2pt]
19g & Outlook and wider use & yes & Discussion (Retrieval comes first, Evidence that existed before the review, Design lessons). Conclusion (the planned physician grading and the regular refresh of the item pool). Protocol of the physician validation: Section~\ref{sec:validation} (Planned physician validation). Limits on generalisation: Limitations (Scale, Task). \\
\end{longtable}
\end{footnotesize}

\begin{footnotesize}
\begin{longtable}{@{}p{0.9cm}p{4.6cm}p{9.5cm}@{}}
\caption{CHART checklist for this report. The guideline is written for chatbot studies, so ``prompts'' is read here as the questions together with the fixed agent and judge prompts. Topics are short labels in our own words. A capitalised name given with a section is one of its paragraph headings.}\label{tab:chart} \\
\toprule
Item & Topic & Where reported \\
\midrule
\endfirsthead
\multicolumn{3}{@{}l}{Table~\ref{tab:chart} (continued)} \\[2pt]
\toprule
Item & Topic & Where reported \\
\midrule
\endhead
\bottomrule
\endfoot
\multicolumn{3}{@{}l}{\emph{Title and abstract}} \\*
1a & Evaluation named in the title & Title, which names the evaluation of LLM agents and the clinical evidence task. \\[2pt]
1b & Structured abstract & Not applicable: the Abstract is one unstructured paragraph, the usual form for an arXiv preprint. \\[6pt]
\multicolumn{3}{@{}l}{\emph{Introduction}} \\*
2a & Background and rationale & Introduction (opening paragraphs) and Section~\ref{sec:related}. \\[2pt]
2b & Objectives and comparisons & Introduction (the study questions and the list of contributions). Intended readers of the answers: Introduction (opening paragraph). Agents and conditions compared: Section~\ref{sec:harness}. Outcome measures: Section~\ref{sec:scoring} (Metrics). \\[6pt]
\multicolumn{3}{@{}l}{\emph{Methods}} \\*
3a & Model names, versions, dates & Names, version tags and weight formats: Section~\ref{sec:harness} (Models) and Table~\ref{tab:models}. Not reported: when each model was released or last updated. The Ollama versions of the runs are stated as far as they are known (Limitations, Conditions). \\[2pt]
3b & Open or proprietary models & Section~\ref{sec:harness} (Models): all agents and judges are open-weight models run locally, and no API model was evaluated. \\[2pt]
4a & Type of model & Section~\ref{sec:harness} (Agent, Models): released open-weight models at the quantisation in Table~\ref{tab:models}, with a fixed context window and literature tools supplied by the harness. The models were used as released, without training or fine-tuning (Models). \\[2pt]
4b & Reference for each base model & Table~\ref{tab:models} identifies each model by tag, architecture, parameter count and weight format. Not reported: references to the developers' model cards or papers. For the models, the text cites only the Ollama runtime. \\[2pt]
4c & Details of tuned models & Not applicable: no model was trained or fine-tuned. What the harness adds (tools, limits, context window) is in Section~\ref{sec:harness}. \\[2pt]
5a & History of the prompts & Questions: built from the source reviews (Section~\ref{sec:construction}, Items). Agent and judge prompts: final versions with version labels in Appendix~\ref{app:prompts}. Not reported: the steps by which these prompts reached their final versions, apart from a change after a pilot (Section~\ref{sec:scoring}, Judge) and one after the empty-answer check (Section~\ref{sec:rubrics}, Drafting and checks). \\[2pt]
5ai & Origin of the prompts & Section~\ref{sec:construction} (Source reviews, Items): each question is built from the primary comparison and outcome of a Cochrane review. \\[2pt]
5aii & People who wrote the prompts & The single author (Declarations, Author contributions), with AI assistance as described in Declarations (Use of AI assistance). No physician decision enters this version; the physician review of the rubrics is under way (Section~\ref{sec:rubrics}, Physician review). \\[2pt]
5aiii & Public role in prompt writing & Not applicable: patients and the public were not involved, as Declarations (Ethics) states. \\[2pt]
5b & Full text of the prompts & Agent and judge prompts: Appendix~\ref{app:prompts}. One question with its rubric: Appendix~\ref{app:example}. The questions and the draft rubrics: the repository given in Declarations (Data and code availability). \\[2pt]
6a & How the models were accessed & Section~\ref{sec:harness} (Agent, Models): local inference through Ollama on one laptop, driven by a scripted Inspect harness, with no API model. Network use for the literature search: Section~\ref{sec:localfirst} (Local-first operation). \\[2pt]
6b & When and where queried & Section~\ref{sec:harness} (Models): the dates of the agent runs, of the budget sensitivity run and of the re-judging by the other judges, and the country in which the laptop ran them. Date of the PubMed candidate query: Section~\ref{sec:construction} (Source reviews). \\[2pt]
6c & Session handling & Section~\ref{sec:harness} (Agent, Conditions): each rollout is a separate episode on one question, ended by the submit call or by a limit. The judge receives one criterion in each call (Section~\ref{sec:scoring}, Judge). \\[2pt]
6d & Release of all outputs & Declarations (Data and code availability): redacted transcripts in the repository. Everything specific to the questions of the grading sample, including their transcripts, is withheld until the blinded physician grading is complete, and the raw logs with fetched abstracts are not distributed. \\[2pt]
7a & Reference standard & Section~\ref{sec:construction} (Items): direction, estimate, GRADE certainty, downgrade reasons and harms taken from the source review. Item-specific rubric: Section~\ref{sec:rubrics}, Table~\ref{tab:template} and Appendix~\ref{app:example}. \\[2pt]
7b & Evaluation procedure & Section~\ref{sec:scoring} (Judge, Metrics, Second and third judges) and Figure~\ref{fig:overview}. Planned physician grading: Section~\ref{sec:validation} (Planned physician validation). Failure coding: Section~\ref{sec:coding} (Failure coding). \\[2pt]
7bi & Who evaluated the answers & No human graded an answer in this version. Automated judges: Section~\ref{sec:scoring} (Judge, Second and third judges) and Table~\ref{tab:models}. Failure coding: AI coders with a spot check by the author (Section~\ref{sec:coding}, Failure coding). Physician grading is planned (Section~\ref{sec:validation}, Planned physician validation). \\[2pt]
7bii & Public role in evaluation & Not applicable: patients and the public took no part in the evaluation (Declarations, Ethics). \\[2pt]
7biii & Blinding to model identity & The judge input has no model name (Section~\ref{sec:scoring}, Judge, with the prompt in Appendix~\ref{app:prompts}). The failure coding was blind to the agent, the score and the judge's decisions (Section~\ref{sec:coding}, Failure coding). The planned physician grading is blind to model and judge (Section~\ref{sec:validation}, Planned physician validation). \\[2pt]
8 & Basis for the sample size & Not reported: a formal sample-size calculation. Items: every eligible Cochrane review in the stated publication window (Section~\ref{sec:construction}, Source reviews). Precision of the design: the paired minimal detectable difference used for ranking (Section~\ref{sec:scoring}, Metrics). Physician sample and failure-coding sample: Section~\ref{sec:scoring} (Planned physician validation, Failure coding). \\[2pt]
9a & Statistics and repeatability & Section~\ref{sec:scoring} (Metrics): cluster bootstrap over reviews, paired differences, the ranking rule, and reliability over repeated rollouts. Agreement statistics: Section~\ref{sec:validation} (Planned physician validation) and Table~\ref{tab:judges}. Repeated rollouts and the budget sensitivity run: Section~\ref{sec:harness} (Conditions) and Section~\ref{sec:results} (Budget sensitivity). \\[2pt]
9ai & Scoring measures & Section~\ref{sec:rubrics}, with Table~\ref{tab:template} and the score formula. Section~\ref{sec:scoring} (Metrics). \\[6pt]
\multicolumn{3}{@{}l}{\emph{Results}} \\*
10a & Agreement with the reference & Section~\ref{sec:results}: Tables~\ref{tab:open}, \ref{tab:criteria}, \ref{tab:cleanroom}, \ref{tab:budget} and~\ref{tab:splits}, Figures~\ref{fig:conditions} and~\ref{fig:budget}. \\[2pt]
10b & Kinds of deviation & Section~\ref{sec:results} (Where answers lose points) with Table~\ref{tab:criteria}. Section~\ref{sec:failures} with Tables~\ref{tab:access} and~\ref{tab:failures}, Figure~\ref{fig:failures} and examples from the transcripts. \\[2pt]
10c & Safety and bias of answers & Misleading or unsafe content is scored through the penalties for unretrieved citations, irrelevant sources, overclaiming and individual advice (Table~\ref{tab:template}). Rates: Tables~\ref{tab:criteria} and~\ref{tab:audit}. Direction of the wording errors: Section~\ref{sec:failures} and Figure~\ref{fig:failures}. Not reported: a separate assessment of bias in the answers. \\[6pt]
\multicolumn{3}{@{}l}{\emph{Discussion}} \\*
11a & Results against earlier work & Discussion, with comparisons to earlier benchmarks, and Section~\ref{sec:related}. \\[2pt]
11b & Strengths and weaknesses & Limitations and ``What this benchmark does not claim''. Strengths: Discussion (Design lessons, Local-first evaluation). \\[2pt]
11c & Implications & Research and agent design: Discussion (Retrieval comes first, Means hide reliability, Evidence that existed before the review, Local-first evaluation). Regulation: Section~\ref{sec:related} (Regulatory context). Limits on use in practice: ``What this benchmark does not claim''. Education is not discussed. \\[6pt]
\multicolumn{3}{@{}l}{\emph{Declarations}} \\*
12a & Competing interests & Declarations (Competing interests). \\[2pt]
12b & Funding & Declarations (Funding). \\[2pt]
12c & Ethics review & Declarations (Ethics): the study uses no patient data and involves no patients, and physicians take part as collaborators, not as research participants. Ethics committee review was not sought, as the same paragraph states. \\[2pt]
12ci & Protection of patient data & Not applicable: no patient data are used (Declarations, Ethics). Section~\ref{sec:localfirst} (Local-first operation) notes that the question text leaves the machine during the literature search. \\[2pt]
12cii & Rights to source material & Section~\ref{sec:construction} (What is stored): no Cochrane abstract or summary text is stored or redistributed, and fetched abstracts are removed from the released logs. Licences of the author's own material: Declarations (Licences). Not reported: whether permission was sought from the publisher of the reviews. \\[2pt]
12d & Protocol & Not reported: a protocol for the whole study. The protocol of the planned physician validation is in the repository (Section~\ref{sec:validation}, Planned physician validation). What else was fixed in advance, with the hash of the frozen sample manifest: Declarations (Registration and pre-specification). \\[2pt]
12e & Access to data, code, weights & Declarations (Data and code availability, Licences): repository, release tag and archive DOI. Model weights: open-weight models identified by the tags in Table~\ref{tab:models}. \\
\end{longtable}
\end{footnotesize}

\end{document}